\documentclass[lettersize,journal]{IEEEtran}
\usepackage{amsmath,amsfonts}
\usepackage{algorithm}
\usepackage{array}
\usepackage[caption=false, position=top, font=normalsize,labelfont=sf,textfont=sf]{subfig}
\usepackage{textcomp}
\usepackage{stfloats}
\usepackage{url}
\usepackage{verbatim}
\usepackage{multirow}
\usepackage{graphicx}
\usepackage{cite}
\usepackage{times}
\usepackage{epsfig}
\usepackage{amssymb}
\usepackage{makecell}
\usepackage{algpseudocode}
\usepackage{caption}
\usepackage{times}
\usepackage{amsmath}

\usepackage[breaklinks=true,bookmarks=false]{hyperref}
\begin{document}

%%%%%%%%% TITLE
\title{VisIRNet: Deep Image Alignment for UAV-taken Visible and Infrared Image Pairs}

\author{ Sedat Özer, Alain P. Ndigande %
\thanks{Both authors are with the Ozer Lab, Dept. of Computer Science, Ozyegin University, Istanbul, Turkiye. Contact email: sedatist@gmail.com}%
\thanks{Manuscript is received on August 04, 2023.}
\thanks{This is the authors' version. This work is accepted for publication at IEEE Transactions on Geoscience and Remote Sensing. }
}

% <-this % stops a space
% \thanks{Manuscript received April 19, 2021.}}

% \markboth{Journal of \LaTeX\ Class Files,~Vol.~14, No.~8, August~2021}% %{Shell \MakeLowercase{\textit{et al.}}: A Sample Article Using IEEEtran.cls for IEEE Journals}

%  \IEEEpubid{0000--0000/00\$00.00~\copyright~2021 IEEE}

\maketitle
%\thispagestyle{empty}

%%%%%%%%% ABSTRACT
\begin{abstract}
   This paper proposes a deep learning based solution for multi-modal image alignment regarding UAV-taken images. Many recently proposed state-of-the-art alignment techniques rely on using Lucas-Kanade (LK) based solutions for a successful alignment. However, we show that we can achieve state of the art results without using LK-based methods. Our approach carefully utilizes a two-branch based convolutional neural network (CNN) based on feature embedding blocks. We propose two variants of our approach, where in the first  variant (ModelA), we directly predict the new coordinates of only the four corners of the image to be aligned; and in the second one (ModelB), we predict the homography matrix directly. Applying alignment on the image corners forces algorithm to match only those four corners as opposed to computing and matching many (key)points, since the latter may cause many outliers, yielding less accurate alignment. We test our proposed approach on four aerial datasets and obtain state of the art results, when compared to the existing recent deep LK-based architectures. 
\end{abstract}


%%%%%%%%% BODY TEXT
\section{Introduction}
Recent advancements in Unmanned Aerial Vehicle (UAV) technologies, computing and sensor technologies, allowed use of UAVs for various earth observation applications. Many UAV systems are equipped with multiple cameras today, as cameras provide reasonable and relatively reliable information about the surrounding scene in the form of multiple images or image pairs. Such image pairs can be taken by different cameras, at different view-points, different modalities or at different resolutions. In such situations, the same objects or the same features might appear at different coordinates on each image and, therefore, an image alignment (registration) step is needed prior to applying many other image based computer vision applications such as image fusion, object detection, segmentation or object tracking as in \cite{ozer2022siamesefuse,ozkanouglu2022infragan,ozer2023offloading}.

The infrared spectrum and visible spectrum may reflect different properties of the same scene. Consequently, images taken in those modalities, typically, differ from each other. On many digital cameras, the visible spectrum is captured and stored in the form of Red-Green-Blue (RGB) image model and a typical visible spectrum camera captures visible light ranging from approximately 400 nanometers to 700 nanometers in wavelength \cite{visibleSpectrum, s20123492,9323397}. Infrared cameras, on the other hand, capture wavelengths longer than those of visible light, falling between 700 nanometers and 10000 nanometers \cite{IRVISIBLE}. Infrared images can be further categorized into different wavelength ranges as in near-infrared (NIR), mid-infrared (MIR), and far-infrared (FIR) capturing different types of information in the spectrum \cite{IRVISIBLE, InfraredSpectrums, IRSpectrums, Gade2014}. 

Image alignment is, essentially, the process of mapping the pixel coordinates from different coordinate system(s) into one common coordinate system. This problem is studied under different names including image registration and image alignment. We will also use the terms {\it alignment} and {\it registration} interchangeably in this paper. Typically, alignment is done in the form of image pairs mapping from one image (source) onto the other one (target)~\cite{books/daglib/0028974}. Image alignment is a common problem that exists in many image-based applications where both of the target and source images can be acquired by sensors using the same modality or using different modalities. There is a wide range of applications of image alignment in many fields including medical imaging~\cite{medical,Hill_2001}, UAV applications~\cite{uav,uav2}, image stitching ~\cite{8663915} and remote sensing applications \cite{1499028, Song:10, LIFuyu:14, Bentoutou, 10.1007/978-3-030-51971-1_46}.

\begin{figure}[t!]
% \vspace*{-1cm}
\begin{center}
\includegraphics[width=0.98\linewidth]{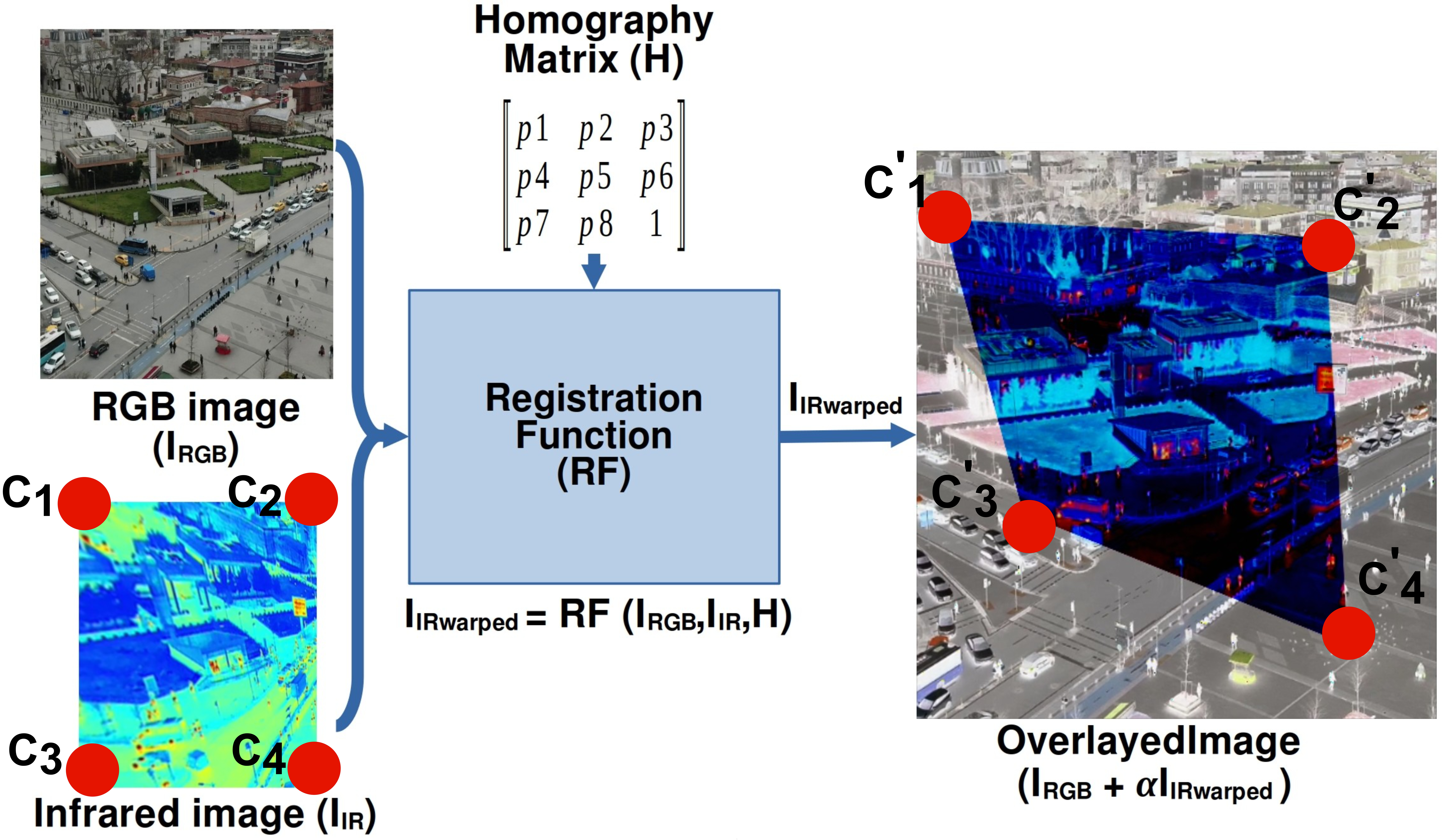}
\end{center} 
   % \vspace*{-6mm}
   \caption{An overview of the image alignment process is shown. On the left, input RGB, $I_{\text{RGB}}$ ({\footnotesize $192\times192$} pixels) and IR, $I_{\text{IR}}$ ({\footnotesize $128\times128$} pixels) images are shown. The $I_{\text{IR}}$ is shown in pseudocolors. Both images are given as input to the registration stage where the transformation parameters represented by the homography matrix ($H$) are predicted. After the registration process, the $I_{\text{IR}}$ is transformed (warped) onto the $I_{\text{RGB}}$ space by locating the positions of $c_1$, $c_2$, $c_3$, and $c_4$ as $c_{1}^{\prime}$, $c_{2}^{\prime}$, $c_{3}^{\prime}$, and $c_4^{\prime}$. The warped $I_{\text{IR}}$ is overlayed (where {\footnotesize $\alpha = 0.4$}) on the $I_{\text{RGB}}$.} 
\label{fig:problemDefinition}
\end{figure}

Image alignment, in many cases, can be reduced to the problem of estimating the parameters of the perspective transformation between two images acquired by two separate cameras, where we assume that the cameras are located on the same UAV system. Fig.~\ref{fig:problemDefinition} summarizes such an image alignment process where the input consists of a higher resolution RGB image (e.g., {\footnotesize $192\times192$} pixels) and a lower resolution IR image (e.g., {\footnotesize $128\times128$} pixels  visualized in pseudocolors in the figure). The output of the registration algorithm is the registered (aligned) IR image on the RGB image's coordinate system. As perspective transformation \cite{hartley2003multiple} is typically enough for UAV setups containing nearby onboard cameras, our registration process uses a registration function based on the Homography (\textbf{H}) matrix. \textbf{H} contains 8 unknown (projection) parameters and the goal of the registration process is predicting those 8 unknown parameters, directly or indirectly.

In the relevant literature, registering RGB and IR image pairs is done by using both classical techniques (such as Scale-Invariant Feature Transform, SIFT,~\cite{lowe2004distinctive} along with the Random Sample Consensus, RANSAC,~\cite{fischler1981random} algorithm as in \cite{siftRansac}) and by using more recent deep learning based techniques as in \cite{electronics12040788, Chang_2017_CVPR, DBLP:journals/corr/abs-2104-11693}. Classical techniques include feature-based~\cite{875,8712669} and intensity-based ~\cite{pixelB} methods. Feature-based~\cite{875,8712669} methods essentially find correspondences between the detected salient features from images~\cite{article0012}. Salient features are computed by using approaches such as SIFT ~\cite{lowe1999object}, Speeded-Up Robust Features (SURF)~\cite{bay2006surf},  Harris Corner~\cite{Harris88alvey}, Shi-Tomas corner detectors~\cite{shiTomasi} in each image. The features from both images are then matched to find the correspondences as in \cite{ransac,ozerfeaturematching,ozer2018similarity}, and to compute the transformation parameters in the form of homography matrix. The RANSAC~\cite{ransac} algorithm is commonly used to compute the homography matrix that minimizes the total number of outliers in the literature. Intensity-based ~\cite{pixelB} methods compare intensity patterns in images via similarity metrics. By estimating the movement of each pixel, optical flow is computed and used to represent the overall motion parameters. In ~\cite{lk1,baker2004lucas} uses LK based algorithms that take the initial parameters and iteratively estimate a small change in the parameters to minimize the error. A typical intensity-based registration technique, essentially uses a form of similarity as its metric or as its registration criteria including Mean Squared Error (MSE)~\cite{gonzalez2008digital}, cross-correlation~\cite{lewis1995fast}, Structural Similarity Index (SSIM) and Peak Signal-to-Noise Ratio (PSNR)~\cite{wang2004image}. Such metrics are not sufficient when source image and target image are acquired by different modalities. This can yield poor performance when such intensity based method are used.

Overall, such major classical approaches, typically, are based on finding and matching similar salient keypoints in image pairs, and therefore, they can yield unsatisfactory results in various multi-modal registration applications.

Relevant deep alignment approaches are using a form of keypoint matching, template matching or Lukas-Kanade (LK) based approaches as in \cite{lei2021deep,Chang_2017_CVPR}. Those techniques typically consider multiple points or important regions in images to compute the homography matrix \textbf{H} which contains the transformation parameters. However, having the information of four matching points represented by their corresponding 2D coordinates $(x_i,y_i)$, where $i=1,2,3,4$ is sufficient to estimate \textbf{H}. Therefore, if found accurately, four matching image-corner points between the IR and RGB images would be enough to perform accurate registration between the IR and RGB images. While many techniques based on keypoint extraction can be employed to find matching keypoints between the images, we argue that the corner points on the borders of one image can also be considered as keypoints, and by using those corners of the image, we do not need to utilize any keypoint extraction step.

In this paper, we propose a novel deep approach for registering IR and RGB image pairs, where instead of predicting the homography matrix directly, we predict the location of the four corner points of the entire image directly. This approach removes the additional iterative steps introduced by LK based algorithms and eliminates the steps of computing and finding important keypoints. Our main contributions can be listed as follows:
(i)  we introduce a novel deep approach for alignment problems of IR images onto RGB images taken by UAVs, where the resolutions of the input images differ from each other;
(ii)  we introduce a novel two-branch based deep solution for registration without relying on the Lukas-Kanade based iterative methods;
(iii) instead of predicting the homography matrix directly, we predict the corresponding coordinates of the four corner points of the smaller image on the larger image;
(iv) we study and report the performance of our approach on multiple aerial datasets and present the state of the art results.

%--------------------------------------
%%%%%%%%%%%%%%%%%%%%%%%%%%%%%%%%%%%%%%%%%%%%%%% RELATED WORKS %%%%%%%%%%%%%%%%%%%%%%%%%%%%%%%%%%%%%%%%%%%%%%%%%%%%%%%%%%%
\section{Related Work}
\label{RelatedWork}

Many recent techniques performing image alignment rely on deep learning. Convolutional Neural Networks (CNNs) form a pipeline of convolutional layers where filters learn unique features at distinct levels of the network. For example, the authors of \cite{DBLP:journals/corr/DeToneMR16}  proposed a Deep Image Homography Estimation Network (DHN) that uses CNNs to learn meaningful features in both images and it directly predicts the eight affine transformation parameters of the homography matrix. Later, the authors of \cite{Le_2020_CVPR} proposed using a series of networks to regress the homograph parameters
in their approach. The latter networks in their proposed architecture aims to gradually improve the performance of the earlier networks. Their method builds on top of DHN~\cite{DBLP:journals/corr/DeToneMR16}. Another work in~\cite{Chang_2017_CVPR}, proposed incorporating the LK algorithm in the deep learning pipeline. 

The authors of~\cite{DBLP:journals/corr/abs-2104-11693} used a CNN based network and introduced a learning based Lucas-Kanade block. In their work, they designed modality specific pipelines for both source and  template images, respectively. At the end of each block, there is a unique feature construction function. Instead of using direct output feature maps, they constructed features based on Eigen values and Eigen vectors of the output feature maps. The features constructed from the source and template network channels have a similar learned representation. Transformation parameters found at a lower scale are given as in input to the next level and the LK algorithm iterates until a certain threshold is reached. In another work, the authors of ~\cite{2023ITIP...32.1078D} utilized disentangled convolutional sparse coding to separate domain-specific and shared features of multi-modal images for improved accuracy of registration. Multi-scale Generative Adversarial Networks (GANs) are also used to estimate homography parameters as in ~\cite{electronics12040788}.

%%%%%%%%%%%%%%%%%%%%%%%%%%%%%%%%%
\begin{figure}[t!]
% \vspace*{-0.38cm}
\begin{center}
\includegraphics[width=.99\linewidth]{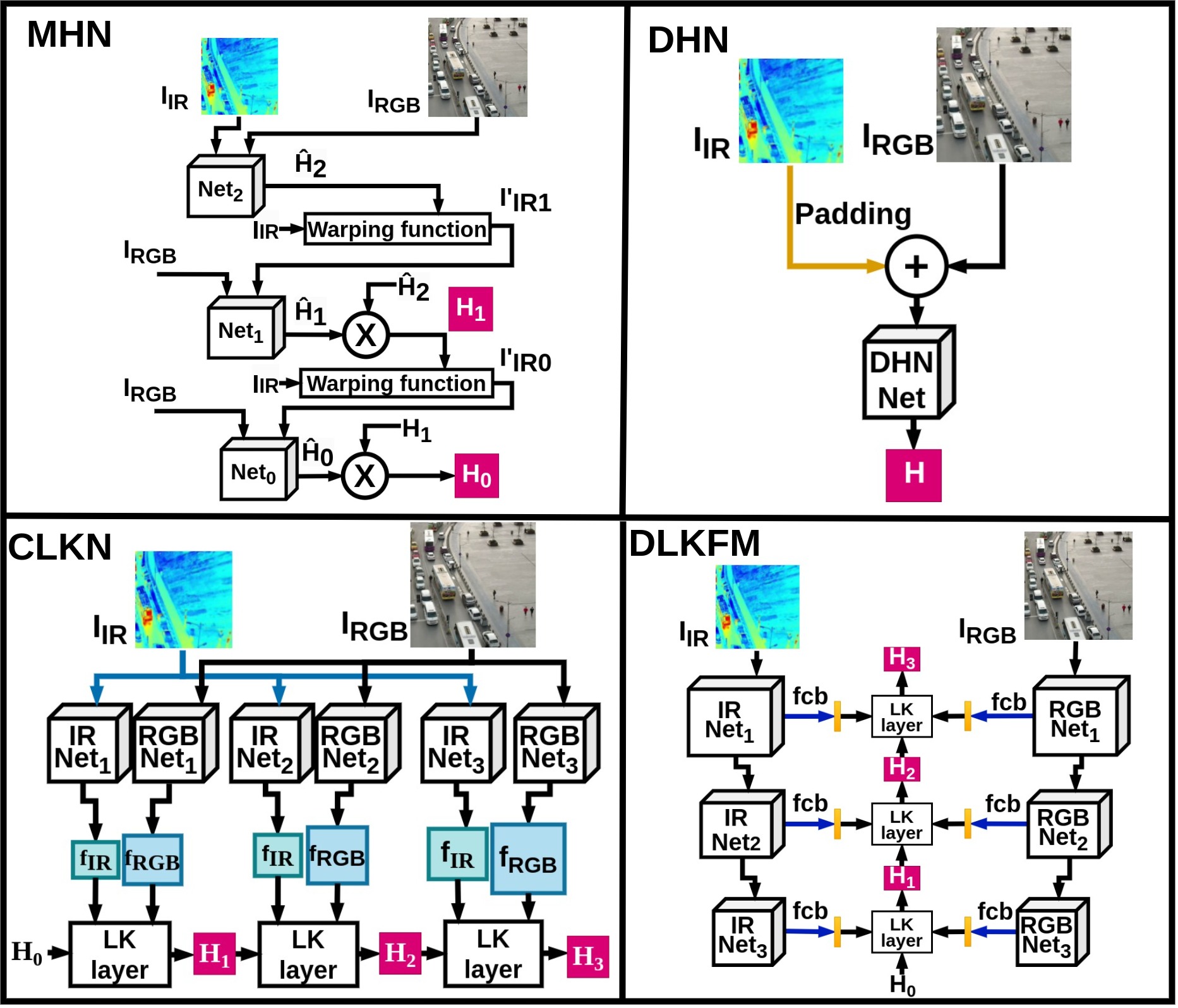}
%\fbox{\rule{0pt}{2in} \rule{.9\linewidth}{0pt}}
\end{center}
   % \vspace*{-6mm}
   \caption{In this figure, we summarize the architectures of various recently proposed deep alignment algorithms including DHN~\cite{DBLP:journals/corr/DeToneMR16}, MHN~\cite{Le_2020_CVPR}, CLKN~\cite{Chang_2017_CVPR} and  DLKFM~\cite{DBLP:journals/corr/abs-2104-11693}. While DHN and  MHN  predict the homography parameters \textbf{H}; CLKN and  DLKFM rely on using Lucas-Kanade (LK) based iterative approach and they use feature maps  at different resolutions. By doing so, they predict homography in steps \textbf{H$_{\textbf{i}}$} where each step aims to correct the previous prediction.
   }
   
\label{fig:WorksSummarizedStructures}
% \vspace*{-6mm}
\end{figure}

The architectural comparisons of the above-mentioned multiple networks is provided in Fig.\ref{fig:WorksSummarizedStructures}.  In DHN~\cite{DBLP:journals/corr/DeToneMR16}, the image to be transformed (it is noted as I$_{IR}$ in the figure) is padded to have the same dimensions as the target image (I$_{RGB}$) and they are concatenated channel-wise. The concatenated images are given to the Deep Homography Network (DHN) for the direct regression of the 8 values of the homography matrix. On the other hand, Multi-scale homography estimation (MHN)~\cite{Le_2020_CVPR} adapts using a series of networks (Net$_{i}$). The inputs for Net$_{2}$ are a concatenation of I$_{IR}$ and I$_{RGB}$. For the succeeding levels, first, the warping function performs the projective inverse warping operation on the infrared image (I$_{IR}$) via the homography matrix which was predicted at the previous level. The resulting image (I$^{'}_{IR_{i}}$) is first concatenated with I$_{RGB}$ and then given as input to the Net$_{i}$. For the following levels, the current  matrix and previously predicted matrices are multiplied to form the final prediction. This way MHN aims to learn correcting mistakes made in the earlier levels. Cascaded Lucas-Kanade Network (CLKN)~\cite{Chang_2017_CVPR} uses separate networks for each modality. They use levels of different scales in the form of feature pyramid networks, and perform registration from smallest to the largest. The homography matrix from the earlier LK-layer is given as input to the next. Deep Lucas-Kanade Feature Maps (DLKFM)~\cite{DBLP:journals/corr/abs-2104-11693} also performs coarse to fine registration as shown in Fig.\ref{fig:WorksSummarizedStructures}. It uses a special feature construction block called (\textbf{fcb}). The (\textbf{fcb}) block takes in the feature maps and transforms them into new feature based on the Eigen vectors and covariance matrix. The constructed features capture principal information and the registration is performed on the constructed feature maps, thus, it aims to increase the accuracy of the LK-layer. Our approach uses separate feature embedding blocks to process each modality separately. It is trained to extract modality specific features so that the output feature maps of different modalities can have similar feature representations.

%---------------------------------------------------
%%%%%%%%%%%%%%%%%%%%%%%%%% PROPOSED METHOD %%%%%%%%%%%%%%%%%%%%%%%%%%%%%
\section{Proposed Approach: VisIRNet}
\label{Approach}
In our proposed approach, we aim at performing accurate, single and multi-modal image registration which is free of the iterative nature of LK-based algorithms. We name our network VisIRNet, where we aim to predict the location of the corners of the input image on the target image directly, since having four matching points is sufficient to compute the homography parameters. In our proposed architecture, we assume that there are two input images with different resolutions. The overview of our architecture is given in Fig.\ref{fig:networkStructure}. Our approach first processes two inputs separately by passing them through their respective feature embedding blocks and extracts representative features. Those features are then combined and given to the regression block as input. The goal of the regression block is computing the transformation parameters accurately. The output of the regression block is eight dimensional (which can represent the total number of homography parameters or the coordinates of the four corner points of the source image on the target image).

\subsection{Preliminaries}
%%%%%%%PRELIMINARIES%%%%%%
\noindent \textbf{Perspective Transformation}: Here, by perspective transformation, we mean a linear transformation in the homogenous coordinate system which, in some sense, warps the source image onto the target image. Homography matrix consists of the transformation parameters needed for the perspective transformation. The elements of the $3\times3$ dimensional homography matrix represent the amount of rotation, translation, scaling and skewing motions. Homography matrix \textbf{H} is defined a follows:
% \vspace*{-3mm}
\begin{equation}
\scriptsize
    \textbf{H} = 
    \begin{bmatrix}
         p_{1} & p_{2}  & p_{3}\\
         p_{4} & p_{5}  & p_{6}\\ 
         p_{7} & p_{8}  & 1
     \end{bmatrix}
\end{equation}
where the last element ($p_{9}$) is set to 1 to ensure the validity of conversion from homogeneous to the cartesian coordinates. \textbf{Warping function} maps a set of coordinates {\scriptsize $[(x_{i},y_{i}),...]$} to another coordinate system via \textbf{H}. Let $c_{i}=(x_{i},y_{i})$ be the location of a point in the coordinates set {\scriptsize $C$} of the source image. Let {\scriptsize $W(c,P)$} be the warping function that warps given coordinate {\small $c$} with parameter set  {\scriptsize $P$} of \textbf{H} to the target image:
\begin{equation}
c_{i}^{'} = W(c_{i},P)
\end{equation}
The warping process is a linear transformation in homogeneous coordinate system. Therefore, the Cartesian coordinates are first transformed into the homogeneous coordinate system by adding the extra $z$ dimension to the 2D Cartesian pixel coordinates. Let $c_{i}$ be the pixel with $x_{i},y_{i}$ coordinates. Homogeneous coordinate of $c_{i}$ can be represented by setting $z-axis$ to 1 i.e., $ c_{h_i} = (x_{i},y_{i},1)$. Once we have the homography matrix, we warp any given $i^{th}$ pixel location $c_{i}$ represented by $(x_{i},y_{i})$ to its warped version $c_{i}^{warped}$ on the other image's Cartesian coordinate as follows;
% \vspace*{-3mm}
\begin{equation}
\scriptsize
    c_{h_i}^{warped} = W(c_{i},P)
    \iff
    \begin{bmatrix}
        x_{i}^{'}\\
        y_{i}^{'}\\ 
        z_{i}^{'}
     \end{bmatrix}
     =
    \begin{bmatrix}
         p_{1} & p_{2}  & p_{3}\\
         p_{4} & p_{5}  & p_{6}\\ 
         p_{7} & p_{8}  & 1
     \end{bmatrix}
     %\times
    \begin{bmatrix}
        x_{i}\\
        y_{i}\\ 
        1
     \end{bmatrix}  
\end{equation}
% \vspace*{-1mm}
where $x_{i}^{'},y_{i}^{'},z_{i}^{'},$ are warped homogeneous coordinates of $c_{i}^{warped}$ which can be converted to Cartesian coordinates by simply division by the $z_{i}^{'}$ value. Therefore, we can obtain the final warped 2D pixel coordinates in Cartesian coordinates as follows:  $c_{i}^{'}=(x_{i}^{'},y_{i}^{'})$, where:
% \vspace*{-3mm}
\begin{equation}
\scriptsize
x_{i}^{warped}=\frac{x_{i}^{'}}{z_{i}^{'}}
\iff
\frac{p_{1} x_{i} + p_{2} y_{i} + p_{3}} {p_{7} x_{i} + p_{8} y_{i} + 1}  
\end{equation}
% and \vspace*{-3mm}
\begin{equation}
\scriptsize
y_{i}^{warped}=\frac{y_{i}^{'}}{z_{i}^{'}}
\iff
\frac{p_{4} x_{i} + p_{5} y_{i} + p_{6}} {p_{7} x_{i} + p_{8} y_{i} + 1}
\end{equation}
% \vspace*{-3mm}

\subsection{Network structure}
\label{lab:networkStructure}
\begin{figure*}[!t]
% \vspace*{-2cm}
\begin{center}
\includegraphics[width=0.85\linewidth]{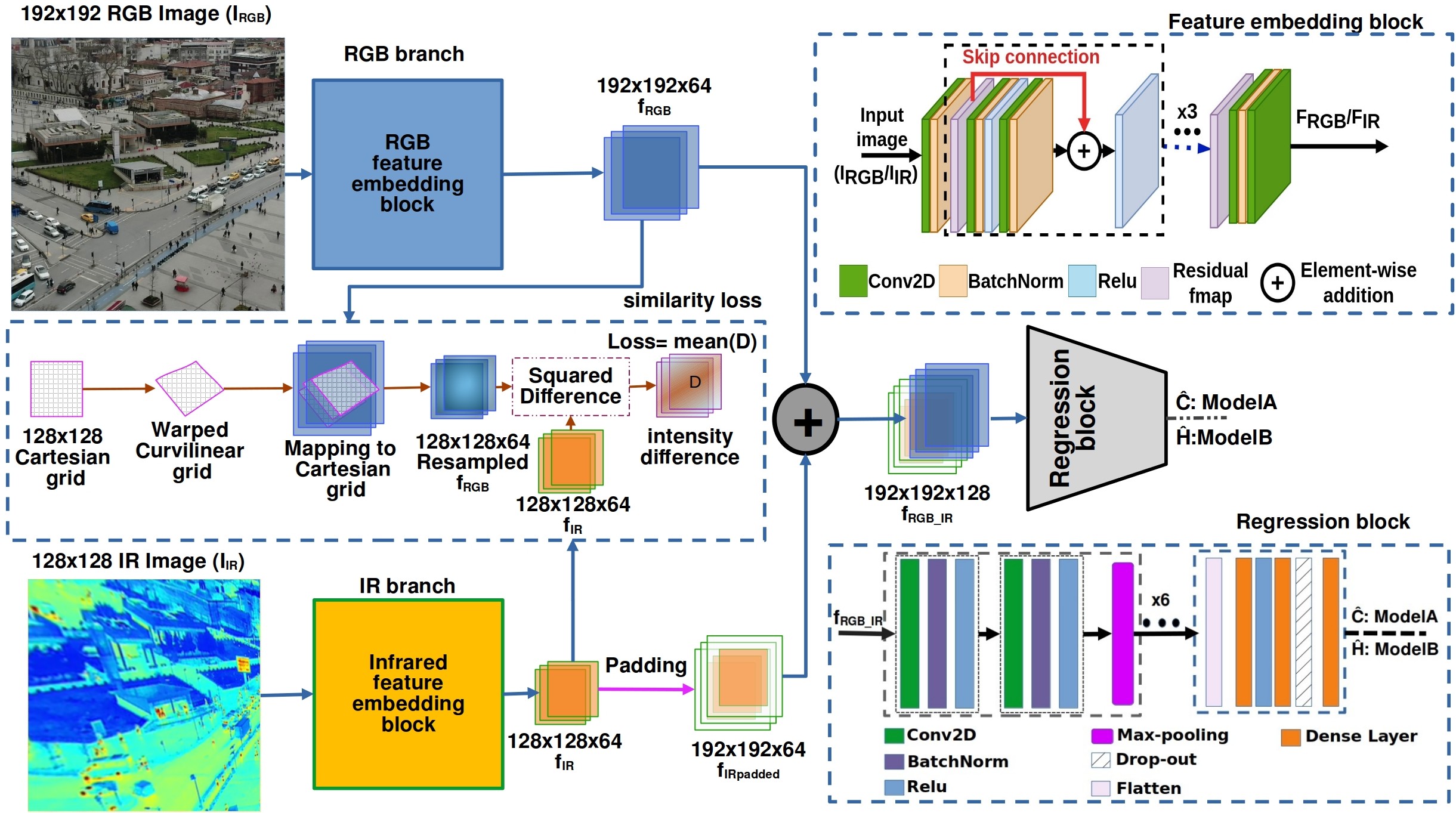}
%\fbox{\rule{0pt}{2in} \rule{.9\linewidth}{0pt}}
\end{center}
   % \vspace*{-6mm}
   \caption{This figure provides an overview of our proposed network architecture. Two parallel branches including RGB branch and IR branch (feature embedding blocks) extract the salient features for RGB and IR images, respectively. Those features are, then channel-wise concatenated and fed into the regression block for direct (ModelB) or indirect (ModelA) homography prediction. I.e., the model can be trained for learning the homography matrix in ModelB or to regress the corresponding coordinates of the four corners of the input IR image on the RGB image in ModelA. The output is 8 dimensional vector (for \textbf{H}), if  ModelB is used; and it is 8 dimensional vector where those 8 values correspond to  the $(x,y)$ coordinates of the 4 corners of the IR image, if modelA is used. The details of the feature embedding block are given on the top corner of the figure (also see Table \ref{table:BackboneStructure}). The details of the regression block are given in the lower right corner of the figure (also see Table \ref{table:NetHeadStructure}).}
\label{fig:networkStructure}
\end{figure*}

 Our proposed network is composed of multi-modal feature embedding blocks (MMFEB) and a regression block (see Fig.\ref{fig:networkStructure}). The regression block is responsible for predicting the 8 homography matrix parameters directly or indirectly. In this paper, we study the performance of two variants of our proposed model and we call them ModelA and ModelB. ModelA predicts the coordinates of the corner points while it's  variant, ModelB, predicts the direct homography parameters. In ModelA, 4 corners are enough to find the homography matrix. Therefore the last layer has 8 neurons for the four $(x,y)$ corner components for the ModelA, or for the eight unknown homography parameters for the modelB.

\label{lab:featureEmbeddingBackbone}
\noindent \textbf{Multi-modal Feature Embedding Backbone}: MMFEB is responsible for producing a combined representative feature set formed of fine level features for both of the input images. The network then, will use that combined representative feature set to transform the source image onto the target image. We adapt the idea of giving RGB and infrared modalities separate branches as in~\cite{DBLP:journals/corr/abs-2104-11693}. We use two identical networks (branches) with same structure but with different parameters for RGB and infrared images, respectively. Therefore, the multi-modal feature embedding block has two parallel branches with identical architectures (however they do not share parameters), namely RGB-branch and infrared branch. We first train the multi-modal feature embedding backbone by using average similarity loss $ \mathcal{L}_{sim}$ (see Eq. \ref{SimilarityLoss}). To compute the similarity loss, we first generate a $128\times128$ recti-linear grid, representing locations in infrared coordinate system as in spatial transformers ~\cite{DBLP:journals/corr/JaderbergSZK15}. Then, we use the ground truth homography matrix to warp the grid onto the RGB coordinate system resulting in a warped curvilinear grid representing projected locations. We use bi-linear interpolation~\cite{ipol.2019.269, Kirkland2010} to sample those warped locations on the RGB feature maps ($f_{RGB}$). After that, we can compute the similarity loss between IR feature maps and re-sampled RGB feature maps. Algorithm \ref{alg:trainingMMFEB} provides the algorithmic details of calculating the similarity loss for the feature embedding block. 

MMFEB is trained by using the $\mathcal{L}_{sim}$ (see Eq. \ref{SimilarityLoss}) which is detailed in Algorithm~\ref{alg:BackboneLossFunction}. Steps for training the MMFEB are given in Algorithm~\ref{alg:trainingMMFEB}. Regression block is trained with homography loss ($\mathcal{L}^{H}_{2}$) in combination with average corner error ( $\mathcal{L}_{Ace}$) (see the subsection "Average Corner Error (Ace)" below for the definitions of $\mathcal{L}_{Ace}$), yielding the total loss { $\mathcal{L}$} to train our model. Table~\ref{table:BackboneStructure} summarises the structure of our used MMFEB.
\begin{table}[!b]
    \scriptsize
    \centering
    \begin{tabular}{l c c c c | c}  \hline
         Layer        &Filter Number            &Filter-dims & Stride    & Padding   &Repetition  \\  \hline
         Conv2D       & 64          & 3x3           & 1         & SAME      &x1     \\
         BN           & -           & -             & -         & -         &      \\  \hline
         Conv2D       & 64          & 3x3           & 1         & SAME      &      \\
         BN           & -           & -             & -         & -         &      \\
         Relu         & -           & -             & -         & -         &x3     \\
         Conv2D       & 64          & 3x3           & 1         & SAME      &      \\
         BN           & -           & -             & -         & -         &      \\
         Relu         & -           & -             & -         & -         &      \\  \hline
         Conv2D       & 64          & 3x3           & 1         & SAME      &      \\
         BN           & -           & -             & -         & -         &x1    \\  
         Conv2D       & 64          & 3x3           & 1         & SAME      &      \\  \hline
    \end{tabular}
    \caption{This table provides the layer-by-layer details of the feature embedding block. There are also skip connections between the layers in this architecture as shown in Fig.\ref{fig:networkStructure}.}
    \label{table:BackboneStructure}
    % \vspace*{-3mm}
\end{table}

%%%%%%%%%%%%%%%%%%%%%%%%%%REGRESSION BLOCK%%%%%%%%%%%%%%%%%%%%%%%%%%%%%%%%%%
\label{lab:regressionBlock}
\noindent \textbf{Regression block}: The second main stage of our pipeline is the regression block which is responsible for making the final prediction. The prediction can be the four corner locations, if ModelA; or the unknown parameters of the homography matrix, if ModelB. $f_{RGB}$ and $f_{IR}$ are the feature maps extracted by passing  the RGB image and infrared image through their respective feature embedding blocks in the feature embedding block. Note that $f_{RGB}$ and $f_{IR}$ have different dimensions. Therefore, we apply zero-padding to the lower dimensional feature maps ($f_{IR}$) so that we can bring its dimensions  to the dimensions of $f_{RGB}$, resulting $f_{IRpadded}$. We concatenate (channel-wise) $f_{IRpadded}$ to  $f_{RGB}$ feature maps coming from infrared and RGB feature embedding blocks and use that as input for the regression block.

The architecture for regression block is further divided in two sub-parts as shown in \textit{Fig.}~\ref{fig:networkStructure}. The first part is composed of 6 levels. Apart from the last level, each level is composed of 2 sub-levels followed by a max-pooling layer. Sub-level is a convolution layer followed by a batch normalization layer followed by a relu activation function. sub-levels \textit{m} and \textit{n}  of a level \textit{l} are identical in terms of the filters used, kernel size, stride and padding used for level \textit{l}. the $6^{th}$ level does not have a max-pooling layer. Second part has two 1024-dense layers with relu as activation function followed by a dropout layer and 8-dense output layer for 8 parameters of homography matrix or corner components. Feature maps from the previous part are flattened and given to the second part where homography matrix parameters or corner components are predicted according to the model used. Table~\ref{table:NetHeadStructure} gives detailed information for the first and the second parts of the regression head.

%%%%%%%%%%%%%%%%%%%%%%% Training & Inference %%%%%%%%%%%%%%%%%%%%%%%%%%

%MMFEB training algorithm

\begin{algorithm}[!b]
\footnotesize
\caption{Training steps of the MMFEB }\label{alg:trainingMMFEB}
\begin{algorithmic}
\State \textbf{Inputs:} $I^{*}_{RGB}, I^{*}_{IR}$ \Comment{* indicates whole training set}
\For{$e \gets 0$ to $epochs$}
    \For{$batch \gets 0$ to $datasetSize/batchsize$}
        \State  $I_{RGB}=I^{*}_{RGB}[batch]$
        \State  $I_{IR}=I^{*}_{IR}[batch]$
        \State  $H \gets groundTruthHomography$
        \State $f_{IR} \gets RGBbranch(I_{RGB})$
        \State $f_{RGB} \gets IRbranch(I_{IR})$
        \State $simLoss \gets \mathcal{L}_{sim}(f_{IR},f_{RGB},H)$
        \State $Backprop(simLoss)$ Using $AdamOptimizer$
    \EndFor
\EndFor
\end{algorithmic}
\end{algorithm}

%%%%%%%%%% HEAD STRUCTURE %%%%%%%%%%

\begin{table}[!b]
    \parbox{.48\linewidth}{
        \scriptsize
        \centering
        \begin{tabular}{l p{1.2cm} c c c c }  \hline
            Level       &Number of filters / Units             &Filter-dims      &Stride       & Padding  &  Activation   \\  \hline
            L1          & \centering 32           & 3x3        & 1           & SAME    &               \\
            max-pool    & \centering -            & 2x2        & 2           & SAME    &               \\
            L2          & \centering 64           & 3x3        & 1           & SAME    &               \\
            max-pool    & \centering -            & 2x2        & 2           & SAME    &               \\
            L3          & \centering 64           & 3x3        & 1           & SAME    &               \\
            max-pool    & \centering -            & 2x2        & 2           & SAME    &               \\
            L4          & \centering 128          & 3x3        & 1           & SAME    &               \\
            max-pool    & \centering -            & 2x2        & 2           & SAME    &               \\
            L5          & \centering 128          & 3x3        & 1           & SAME    &               \\
            max-pool    & \centering -            & 2x2        & 2           & SAME    &               \\
            L6          & \centering 256          & 3x3        & 1           & SAME    &               \\
            
            \hline
            Flatten          &         &           &            &           &               \\
            \hline
            Dense       & \centering 1024         & -         & -           &          &     Relu      \\
            Dense       & \centering 1024         & -         & -           &          &     Linear    \\
            Dropout     & \centering 20\%         & -         & -           &          &      -        \\
            Dense       & \centering 8            & -         & -           &          &     Linear    \\
            \hline
        \end{tabular}
    }
    \caption{This table provides the layer-by-layer details of the regression block as shown in Figure~\ref{fig:networkStructure}. Levels indicated by L are groups of {\scriptsize $conv2D$ + $BatchNormalization$ + $Relu$}. The {\scriptsize $Conv2D$} layers in each level have the same characteristics and filter dimensions. The number of used filters increases as we get deeper in the architecture.}
     \label{table:NetHeadStructure}
\end{table}

%transformation block training algorithm
\begin{algorithm}[!t]
\footnotesize
\caption{Training step of the regression block }\label{alg:traningTransformationHead}
\begin{algorithmic}
\State let \textbf{M} be RegressionBlock
\State \textbf{Inputs:} $I^{*}_{RGB}, I^{*}_{IR}$ \Comment{* indicates whole training set}
\For{$e \gets 0$ to $epochs$}
    \For{$batch \gets 0$ to $datasetSize/batchsize$}
    
        \State  $I_{RGB}=I^{*}_{RGB}[batch]$
        \State  $I_{IR}=I^{*}_{IR}[batch]$
        \State  $H \gets groundTruthHomography$
        \State $f_{IR} \gets RGBbranch(I_{RGB})$
        \State $f_{RGB} \gets IRbranch(I_{IR})$

        \Ensure $f_{RGB}.shape = 192\times192\times64$
        \Ensure $f_{IR}.shape = 128\times128\times64$

        \State $f_{IRpadded} = zeroPadd(f_{IR})$

        \State $f_{RGB\_IR} = concat(f_{RGB}, f_{IRpadded})$
        \State $\hat{H} = \textbf{M}(f_{RGB\_IR})$
        \State $Loss = \mathcal{L}(\hat{H},H) $
        \State $Backprop(Loss)$ using $AdamOptimizer$ 
    \EndFor
\EndFor
\end{algorithmic}
\end{algorithm}

\label{backbonelossfunction}
\subsection{Loss}
While MMFEB uses \textit{similarity loss}, we used two loss terms  based on the corner error and homography for the regression head.

\noindent \textbf{Similarity loss}: The similarity loss is used to train MMFEB and is defined as follows:
% \vspace*{-2mm}
\begin{equation}\label{SimilarityLoss}
\small 
\mathcal{L}_{sim}=\frac{1}{x*y} \sum_{x=0}^{n} \sum_{y=0}^{n} ({f^{'}_{RGB}(x,y)- f_{IR}(x,y)})^2  
\end{equation}
where, $f_{IR/RGB}(x,y)$ is the value at $(x, y)$ location for respective image feature maps. $f^{'}_{RGB}(x,y)$  is the value at $(x, y)$ location on the re-sampled RGB feature maps. Note that the $(x, y)$ is a location on the coordinate system constrained by the infrared image height and width. The algorithmic details of the  similarity loss are provided in Algorithm \ref{alg:BackboneLossFunction}.

\begin{algorithm}[!t]
\footnotesize
\label{Algorithm3}
\caption{Computing the $\mathcal{L}_{sim}$ loss}\label{alg:BackboneLossFunction}
\begin{algorithmic}
\State $f_{RGB} \gets RGBbranch(I_{RGB})$
\State $f_{IR} \gets IRbranch(I_{IR})$
\State $H \gets groundTruthHomography$
\State $grid_{n\times n} \gets {2x2 grid with Ir dimensions}$
\Ensure $warpedGrid= warpGrid(grid_{n\times n},H^{-1})$
\State $f^{'}_{RGB}= BilinearSampler(f_{RGB},warpedGrid)$
\State $\mathcal{L}_{sim} \gets 0$
\For{$i=0 , i\leq n\times n$}
    \State $Ir_{i}=f_{IR}[i]$
    \State $Rgb_{i}=f^{'}_{RGB}[i]$
    \State $P_{diff} \gets Ir_{i}-Rgb_{i}$
    \State $\mathcal{L}_{sim}\gets  \mathcal{L}_{sim} + P^{2}_{diff}$
\EndFor
\State $\mathcal{L}_{sim} \gets \mathcal{L}_{sim} /  (n\times n) $
\end{algorithmic}
\end{algorithm}

\noindent \textbf{ $ \mathbf{L_2}$ Homography loss term}: ModelB is trained to predict the values of the elements of the homography matrix. Therefore, its output is the 8 elements of a 3x3 matrix (where the ninth element is set to 1). The homography based loss term: ${L}^{H}_{2}$ is defined as follows: let {\small [$p_{i}$ : \textit{(for $i=1,2,3,4,5,6,7,8$), 1]}} be the elements of a {\small $3\times 3$} \textbf{H} ground truth homography matrix. Similarly, let {\small [$\hat{p}_{i}$ : \textit{(for $i=1,2,3,4,5,6,7,8$), 1]}} be elements of  {\small $3\times 3$  \textit{$\hat{H}$}}, the predicted homography matrix. Then, {\small $\mathcal{L}^{H}_{2} = \frac{1}{8} \sum_{i=1}^{8} (p_{i} - \hat{p}_{i})^{2}  $}
where, {\small $\mathcal{L}^{H}_{2}$} represents the homography loss based on the $L_2$ distance.

\noindent \textbf{Average Corner Error (Ace):}
  Ace is computed as the average sum of squared differences between the predicted and ground truth locations of the corner points. For ModelB, we use predicted homography matrix to transform the 4 corners of infrared image onto the coordinate system of RGB image and together with ground truth locations we compute $\mathcal{L}_{Ace}$. Let $e_{i}$ be a corner at the ($x_{i},y_{i}$) coordinates on the infrared image and let $e_{i}^{'}$ be its warped equivalent on the RGB coordinate space such that {\small $e_{i}^{'} = W(e_{i},P)$} where $\textit{W}$ is the warping function.
% \vspace*{-0.30cm}
% \label{Average_Corner_error}
\begin{equation}
\scriptsize
\mathcal{L}_{Ace} = \frac{1}{4} \sum_{i=1}^{4} \textit{D}(e_{i},e_{i}^{'})^{2} =\frac{1}{4} \sum_{i=1}^{4} (W(e_{i},P) - W(e_{i},\hat{P}))^{2}
\end{equation}
where $\textit{D}$ is defined as: {\scriptsize $\textit{D}(e_{i},e_{i}^{'}) = W(e_{i},P) - W(e_{i},\hat{P})$}, and where { $P$} and { $\hat{P}$}  are ground truth and predicted vectorized homography matrices, respectively. The total loss for ModelB, then, is computed as  $\mathcal{L}= \mathcal{L}^{H}_{2}+ \gamma \mathcal{L}Ace$ where $\gamma$ is weight factor (a hyperparameter). 

\begin{table}[t]
\scriptsize
\centering
\setlength\tabcolsep{4.9pt}
\begin{tabular}{|c|c|c|c|c|c|}
\hline
\multirow{2}{*}{} & \multicolumn{2}{c|}{SkyData} & \multicolumn{2}{c|}{VEDAI} & \multirow{2}{*}{Average} \\ \cline{2-5} 
& $\mathcal{L}_{Ace}$ & ${L}^{H}_{2}$ & $\mathcal{L}_{Ace}$ & ${L}^{H}_{2}$ &  \\ \hline
$\mathcal{L}_{sim}$ & 18.6 & 21.6 & 19.1 & \textbf{128.3} & \textbf{36.85}\\ \hline
$\mathcal{L}_{MAE}$ & \textbf{18.5} & 35.8 & \textbf{18.5} & 178.1 & 62.7\\ \hline
% $\mathcal{L}_{SSE}$ & 19.1 & 49.0 & 20.7 & 459.3 & 137.025 \\ \hline
$\mathcal{L}_{SSIM}$ & \textbf{18.5} & \textbf{20.1} & 19.1 & 134.2 & 47.9 \\ \hline
\end{tabular}
\caption{Ablation study on using different combinations of loss functions on two different datasets. The loss functions shown in each row are used for the MMFEB block, and the loss functions shown in each column ($\mathcal{L}_{Ace}$ and ${L}^{H}_{2}$) are used for the regression block in our model. Best results are shown in bold. Ace is the metric used to compute the results for each loss function combination. The last column shows the average Ace value for each loss function used in the MMFEB block. On average, $\mathcal{L}_{sim}$ yielded the best results.  }
\label{tab:ablationstudy}
\end{table}

In ModelB, we predict the $x$ and $y$ locations of the 4 corner points, instead of computing the homography matrix. This makes it possible for the network to learn to predict exact locations (landmarks) instead of focusing on one solution. As shown in our experiments, (see \textit{Fig.}~\ref{fig:QualitativeResults} for qualitative and \textit{Fig.}~\ref{fig:AceResults} for quantitative results), ModelA converges faster and yields better results, while minimizing outliers. 
We use a slightly modified version of  $\mathcal{L}_{Ace}$ for ModelA such that $\hat{e}_{i}$ becomes the ground truth corner coordinate in RGB coordinate space. For ModelA, $\mathcal{L}_{Ace}$ is defined as follows:
% \vspace*{-0.30cm}
\begin{equation}
\footnotesize
\mathcal{L}_{Ace} = \frac{1}{4} \sum_{i=1}^{4} \textit(e_{i} - \hat{e}_{i})^{2}    
\end{equation}

In addition to these loss functions, we also used additional loss functions in the MMFEB block during our ablation study. Those functions are $\mathcal{L}_{MAE}$ and 
$\mathcal{L}_{SSIM}$. They are briefly defined below.
%%%%%%%%%%%% additional loss functions 
% mae
\begin{equation}\label{mMeanAbsoluteError}
\centering
\small 
\mathcal{L}_{MAE}=\frac{1}{x*y} \sum_{x=0}^{n} \sum_{y=0}^{n} |{f^{'}_{RGB}(x,y)- f_{IR}(x,y)}|  
\end{equation}

%%SSIM

\begin{equation}\label{SSIMloss}
\centering
\small 
\mathcal{L}_{SSIM} = 1-SSIM(f^{'}_{RGB}(x,y), f_{IR}(x,y)
\end{equation}
where SSIM is used as also used and defined in \cite{ozer2022siamesefuse}.

\normalsize

\section{Experiments}
In this section, we describe our experimental procedures, used datasets and our metrics. Below we describe our used datasets.
\label{lab:experiments}

\noindent \textbf{Datasets:}
In our experiments we use Skydata\footnote{\url{www.skydatachallenge.com}} containing RGB and IR image pairs, MSCOCO \cite{DBLP:journals/corr/LinMBHPRDZ14}, Google-Maps, and  Google-Earth (as taken from DLKFM ~\cite{DBLP:journals/corr/abs-2104-11693}), VEDAI~\cite{VEDAI} datasets. Please refer Table~\ref{table:DatasetsUsed} for more details about the used datasets in our experiments. %Google-Earth dataset contains RGB images taken at different seasons. 
SkyData is originally a video-based dataset which provides each frame of the videos in image format.

\noindent \textbf{Generating the training and test sets: } 
To train the algorithms we need unregistered and registered (ground truth) image pairs. For SkyData,  we randomly select $m$ frame pairs for each video sequence.

\begin{figure}[t]
\begin{center}
\includegraphics[width=0.74\linewidth]{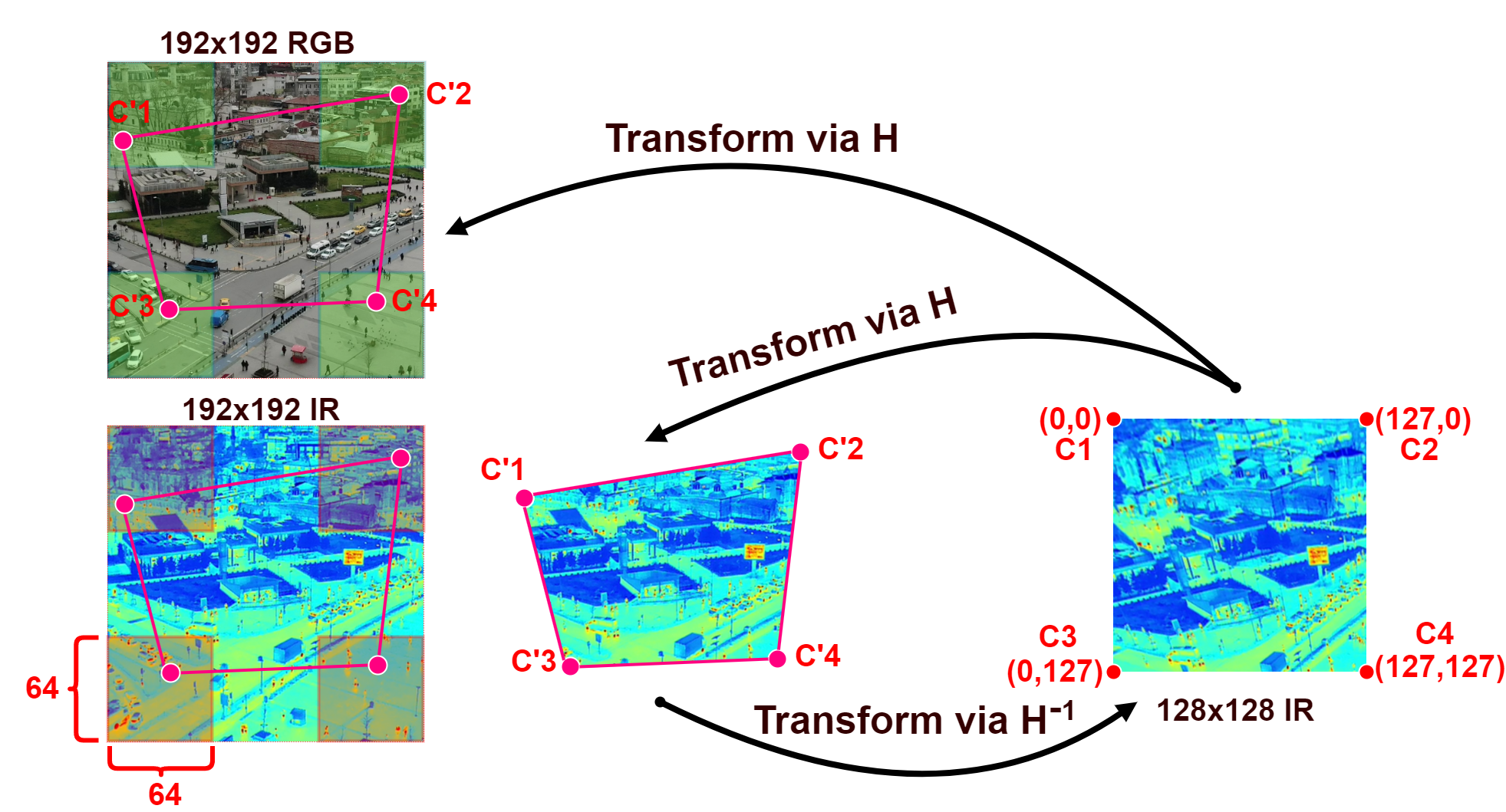}
\end{center}
   % \vspace*{-6mm}
   \caption{This figure shows how to select initial corner points on the registered image pairs and how to generate the training data. First a random image patch is taken from the originally registered IR image. Then the random corners of that patch is transformed into fixed coordinates and after that, the \textbf{H} matrix (and its inverse) performing that transformation is computed.}
\label{fig:TrainingDataGeneration}
% \vspace*{-6mm}
\end{figure}

For each dataset that we use, we  generate the training and test sets as follows:
(i) Select a registered image pair at higher resolutions.
(ii) Sample (crop) regions around the center of the image to get smaller patches of 192x192 pixels. This process is done in parallel for visible and infrared images.
 (iii) If the extracted patches are not sufficiently aligned, manually align them.
 (iv) For each pair, select a subset of the IR image, by randomly selecting 4 distinct locations on the image.
 (v) Find perspective transformation parameters that map those randomly chosen points to the following fixed locations: (0,0) , ($n-1$,0) , ($n-1,n-1$) , (0,$n-1$) so that they can correspond to the corners of the unregistered IR image patch; where we assume that the unregistered $I_{IR}$ is $n\times n$ dimensional (in our experiments $n$ is set to 128). This process creates an unregistered infrared patch (from the already registered ground truth) that needs to be placed back to its true position.
 (vi) Use those 4 initially selected points as the ground truth corners for the registered image. 
 (vii) Repeat process $k$ times to create $k$ different image pairs. This newly-created dataset is then split into training and test sets. 
 (viii) The RGB images are used as the target set and the transformed infrared patches are used as the source set (for both training and testing). 
This process is done on randomly selected registered pairs for each dataset. Fig.\ref{fig:TrainingDataGeneration} also illustrates this process on a pair of RGB and IR images. The list of all the used data sets and their details are summarized in Table~\ref{table:DatasetsUsed}. 

\begin{table}[!t]
\footnotesize
    \centering
    \begin{tabular}{ l l l l }  \hline
        Dataset        & Modality          & Training set  & Test Set  \\ \hline 
        SkyData       & RGB + Infrared    &  27700        & 7990     \\ \hline
        MSCOCO  \cite{DBLP:journals/corr/LinMBHPRDZ14}       & Single modality (RGB)            &  82600        & 6400     \\ \hline
        Google Maps   & RGB + map (vector)       &  8800         & 888      \\ \hline
        Google Earth  & RGB + RGB         &  8750         & 850       \\ \hline
        VEDAI \cite{VEDAI}         & RGB + Infrared    &  8722         & 3738       \\ \hline
    \end{tabular}
    % \vspace{-2mm}
    \caption{\small A summary of the used datasets in our experiments is given. The training and test datasets are generated as explained in Section~\ref{lab:experiments}.}
    \label{table:DatasetsUsed}
\end{table}

% %%%%%%%%% L1 L2 Ace %%%%%%%%%%%%%%%%%
\begin{table}[t]
    \subfloat[t][{\footnotesize Homography loss (${L}^{H}_{2}$)}]{
        \centering
        \begin{minipage}[b]{0.5\textwidth}
            \scriptsize
            \setlength\tabcolsep{.4mm}
            \begin{tabular} {p{1cm} *{1} { p{.78cm} } *{1} { p{.58cm} } p{.7cm}  p{.9cm} *{4}{ p{.8cm}} p{.80cm}}
                \hline
                Model  & BatchS &loss  & mean & std & min & 25\% & 50\% & 75\% & max \\  \hline
                ModelB & \centering 8 & L1  & \textbf{0.6} & 0.44 & 0.03 & 0.32 & \textbf{0.49 }& \textbf{0.76} & 8.38 \\  \hline
                ModelB & \centering 8 & L2  & 1.84 & 4.49 & \textbf{0.0} & 0.37 & 0.89 & 2.03 & 219.33 \\  \hline
                ModelA & \centering 8 & Ace & 2.49 & 4.15 & \textbf{0.0} & 0.48 & 1.23 & 2.85 & 63.65 \\  \hline
                ModelB & \centering 16 & L1 & \textbf{0.6} & \textbf{0.4}   & 0.03 & 0.34 & 0.51 & \textbf{0.76} & \textbf{6.08} \\  \hline
                ModelB & \centering 16 & L2 & 1.94 & 4.05 & \textbf{0.0} & \textbf{0.3} & 0.8 & 2.02 & 92.81 \\  \hline
                ModelA & \centering 16 & Ace & 2.14& 4.1  & \textbf{0.0} & 0.4 & 1.04 & 2.42 & 176.51 \\  \hline
                ModelB & \centering 32 & L1 & 0.7  & 0.46 & 0.02 & 0.4 & 0.61 & 0.9 & 8.17 \\  \hline
                ModelB & \centering 32 & L2 & 2.79 & 5.3  & \textbf{0.0} & 0.5 & 1.31 & 3.08 & 149.63 \\  \hline
                ModelA & \centering 32 & Ace & 2.98& 4.71 & \textbf{0.0} & 0.57 & 1.51 & 3.5 & 81.88 \\  \hline
                ModelB & \centering 64 & L1 & 0.73 & 0.49 & 0.03 & 0.39 & 0.61 & 0.93 & 7.03 \\  \hline
                ModelB & \centering 64 & L2 & 3.62 & 5.94 & \textbf{0.0} & 0.71 & 1.82 & 4.09 & 107.96 \\  \hline
                ModelA & \centering 64 & Ace & 3.65& 5.17 & \textbf{0.0} & 0.91 & 2.22 & 4.57 & 157.6\\ \hline 
            \end{tabular}
        \end{minipage}
    }
    \hfill
    \vspace{.2mm} 
    \subfloat[t][{\footnotesize Average corner error}]{
        \begin{minipage}[b]{0.5\textwidth}
            \scriptsize
            \setlength\tabcolsep{.4mm}
            \begin{tabular} {p{1cm} *{1} { p{.78cm} } *{1} { p{.58cm} } p{.7cm}  p{.9cm} *{4}{ p{.8cm}} p{.80cm}} \\
            \hline
                Model & BatchS & Loss & Mean & Std & Min & 25\% & 50\% & 75\% & Max \\ \hline
                ModelB & \centering 8 & L1 & 21.38 & 44.01 & 2.53 & 8.65 & 12.57 & 18.27 & 235.99 \\ \hline
                ModelB & \centering 8 & L2 & 25.15 & 119.95 & 2.30 & 11.83 & 13.11 & 15.53 & 893.01 \\ \hline
                ModelA & \centering 8 & Ace & 3.96 & \textbf{1.64} & 0.93 & 2.84 & 3.68 & 4.72 & 19.59 \\ \hline
                ModelB & \centering 16 & L1 & 25.83 & 108.53 & 1.66 & 8.02 & 10.97 & 16.27 & 743.54 \\ \hline
                ModelB & \centering 16 & L2 & 31.27 & 143.28 & 2.29 & 10.52 & 13.65 & 18.47 & 1180.64 \\ \hline
                ModelA & \centering 16 & Ace & \textbf{3.83} & 1.77 & \textbf{0.78} & \textbf{2.64} & \textbf{3.46} & \textbf{4.61} & 19.19 \\ \hline
                ModelB & \centering 32 & L1 & 25.17 & 101.33 & 1.87 & 6.08 & 8.80 & 16.97 & 803.25 \\ \hline
                ModelB & \centering 32 & L2 & 6.50 & 4.11 & 2.05 & 5.43 & 6.40 & 7.35 & \textbf{15.07} \\ \hline
                ModelA & \centering 32 & Ace & 5.32 & 1.87 & 1.35 & 4.04 & 5.02 & 6.22 & 22.21 \\ \hline
                ModelB & \centering 64 & L1 & 30.60 & 177.66 & 3.41 & 11.07 & 11.65 & 12.41 & 1298.48 \\ \hline
                ModelB & \centering 64 & L2 & 31.86 & 135.19 & 1.74 & 8.10 & 13.46 & 19.23 & 981.56 \\ \hline
                ModelA & \centering 64 & Ace & 5.01 & 2.01 & 0.95 & 3.62 & 4.65 & 6.0 & 18.94 \\ \hline
            \end{tabular}
        \end{minipage}
    }
    
    \caption{\small A comparison of the results of ModelA and ModelB at various hyperparameters including batchsizes and loss functions. The top table shows the Homography error in (a), while the bottom table shows the results as Average corner error in (b).  Note that the results illustrating low homography error doesn't necessarily imply small registration error. we study the effect of bacthsizes and loss functions. directly predicting homography matrix work but it does not mimimize the registration error as predicting direct corners in our experiments. These results are obtained on Skydata dataset.}
    \label{tab:l12didnotwork}
\end{table}

%-----------------------------------------------Qualitative Results------------------------------------------
\begin{figure*}[t!]
% \vspace*{-2.2cm}
\begin{center}
\includegraphics[width=.94\linewidth,width=.94\linewidth]{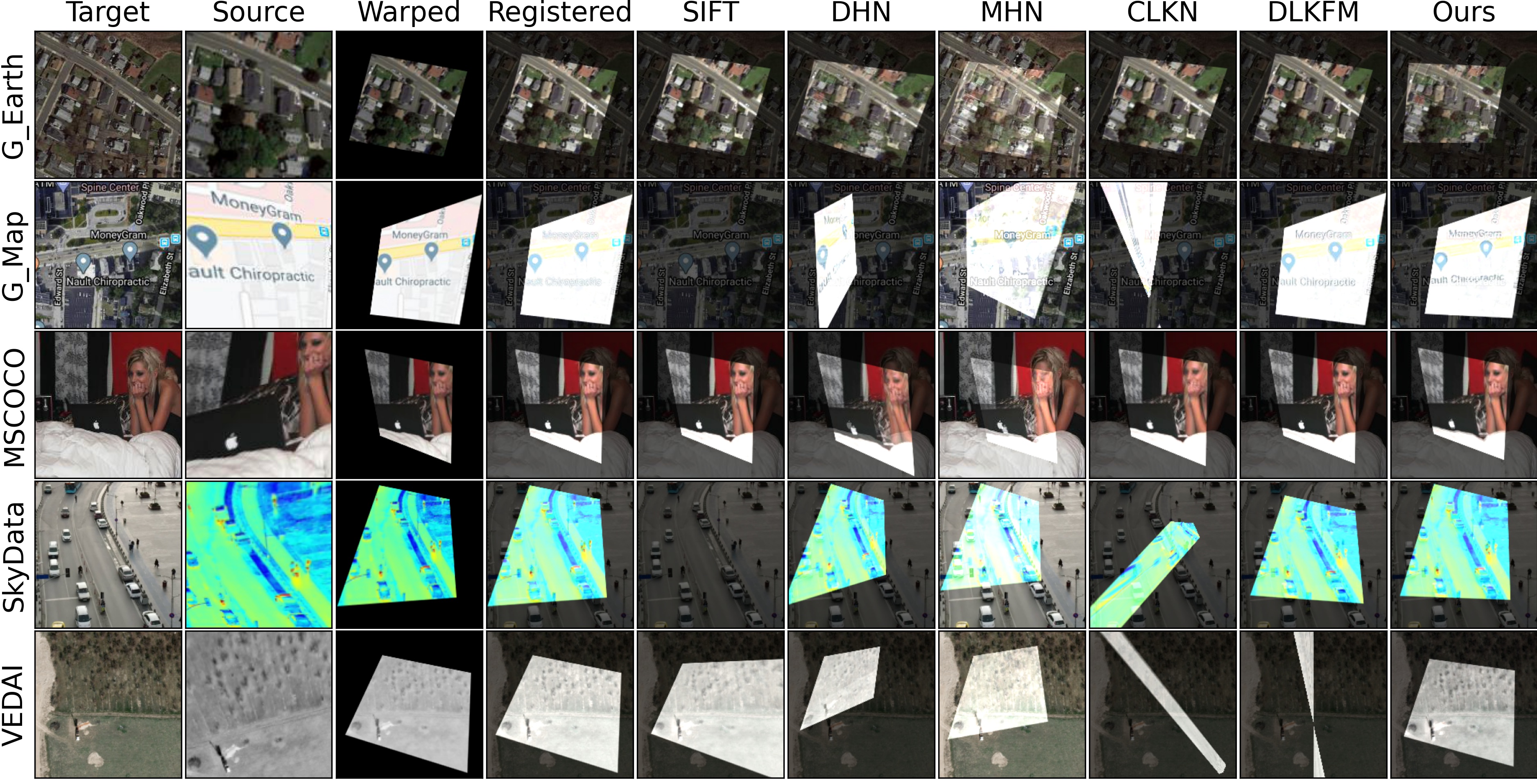}
\end{center}
   % \vspace*{-6mm}
   \caption{This figure shows qualitative results on sample image pairs taken from different datasets. The first two columns show the input image pairs for the algorithms. The target image is $192\times192$ pixels and the source image is $128\times128$ pixels (which covers a scene that is a subset of the target image). The third column shows the ground truth version ($192\times192$ pixels) of the source image on the coordinate system of the target image after being warped. The fourth column shows the ground truth (warped) where the source image is overlayed on the target image ($192\times192$ pixels). The remaining 6 columns show the overlayed results ($192\times192$ pixels), after applying registration with the algorithms in the order of SIFT, DHN, MHN, CLKN, DLKFM and our approach, respectively. Visually, each algorithms' result can be compared to the image in the fourth column.}
\label{fig:QualitativeResults}
% \vspace*{-.7cm}
\end{figure*}

\begin{figure}[t]
  \centering
  \begin{minipage}[b]{0.49\linewidth}
    \centering
    \includegraphics[width=\textwidth]{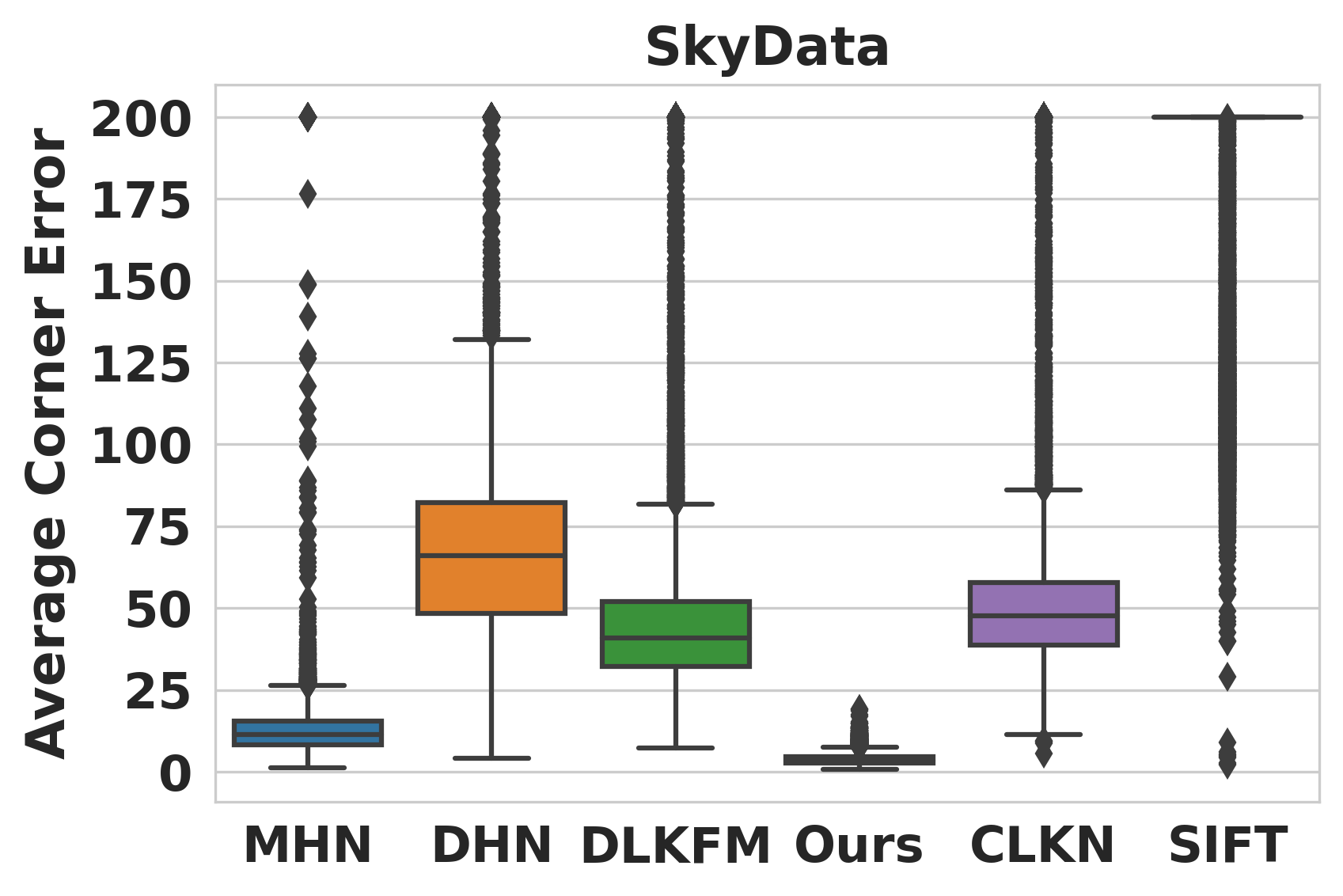}
  \end{minipage}%
  % \quad
  \begin{minipage}[b]{0.49\linewidth}
    \centering
    \includegraphics[width=\textwidth]{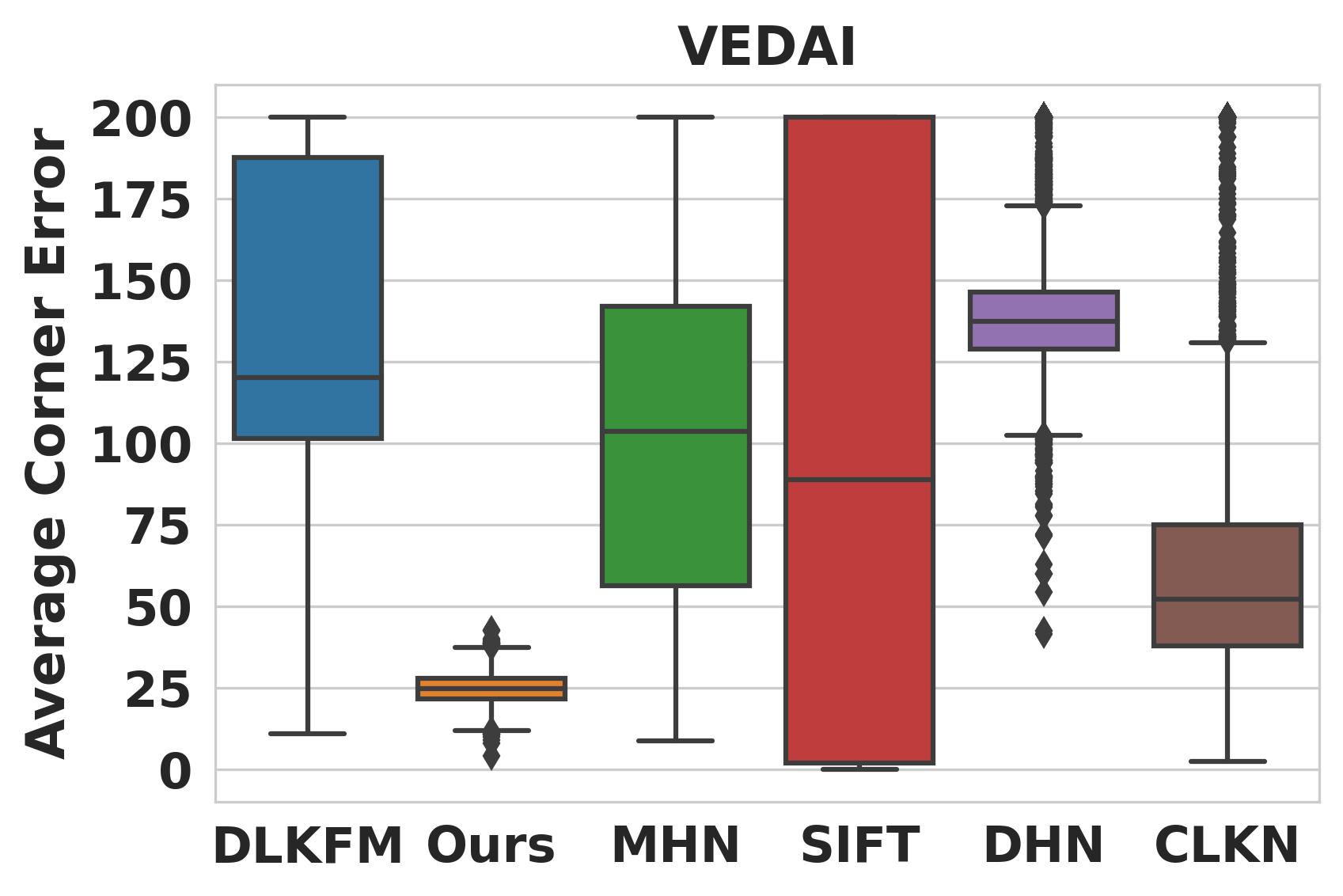}
  \end{minipage}%
   \vspace{1mm}
  \begin{minipage}[b]{0.49\linewidth}
    \centering
    \includegraphics[width=\textwidth]{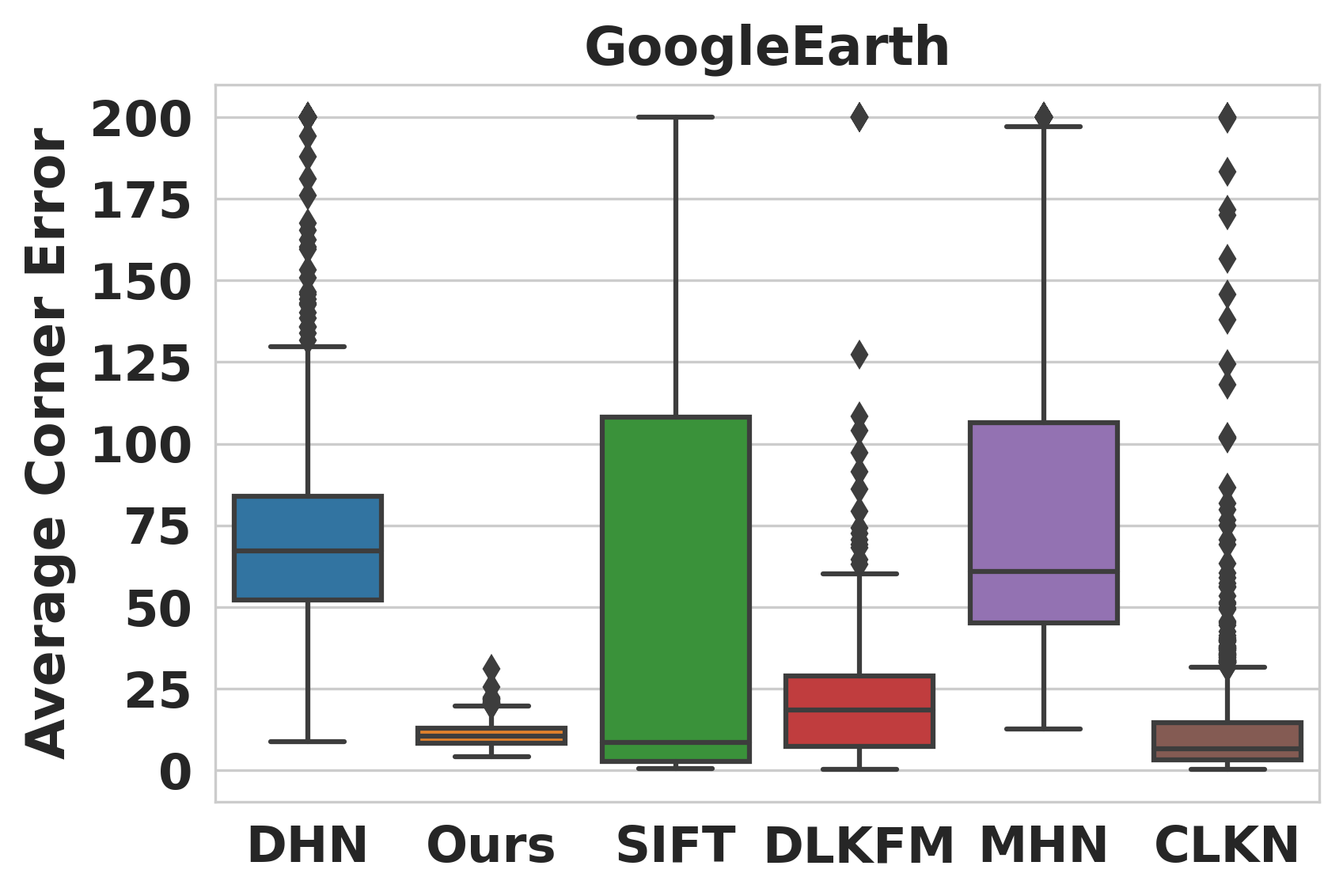}
  \end{minipage}%
  % \quad
  \begin{minipage}[b]{0.49\linewidth}
    \centering
    \includegraphics[width=\textwidth]{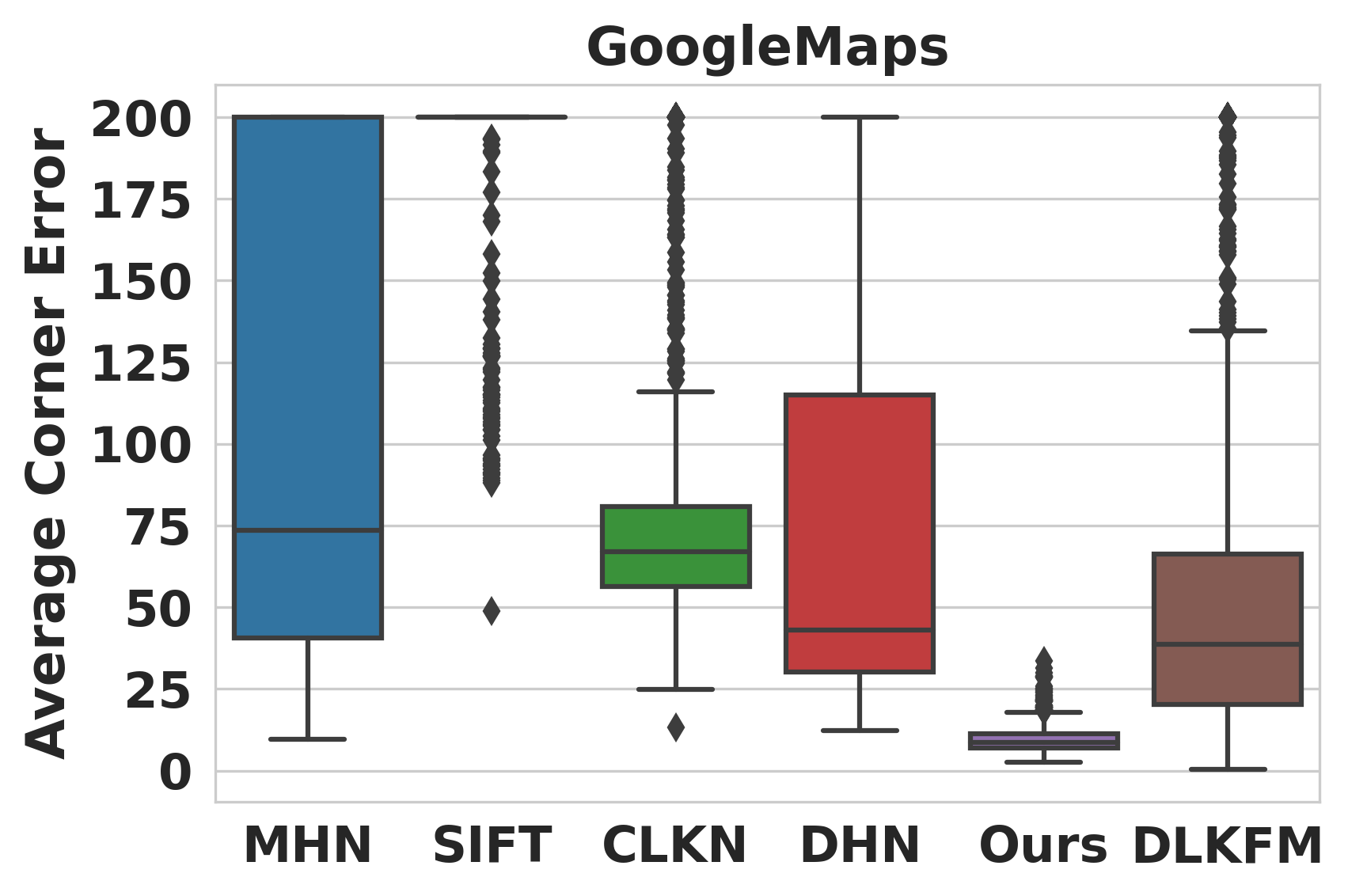}
  \end{minipage}%
   \vspace{1mm} 
     \begin{minipage}[b]{0.49\linewidth}
    \centering
    \includegraphics[width=\textwidth]{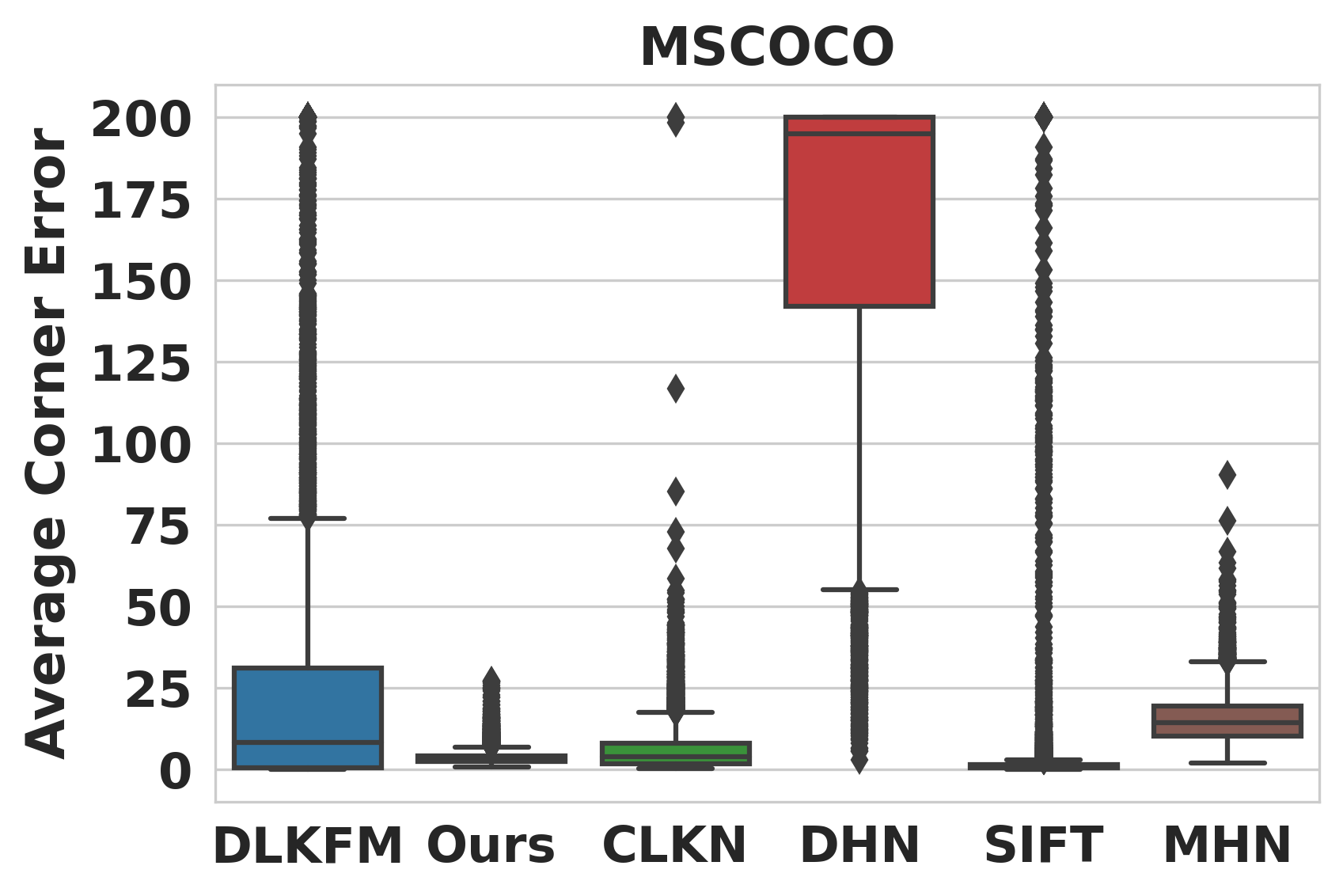}
  \end{minipage}%
  \begin{minipage}[b]{0.49\linewidth}
    \centering
    % \caption*{Legend}
    \includegraphics[width=\textwidth]{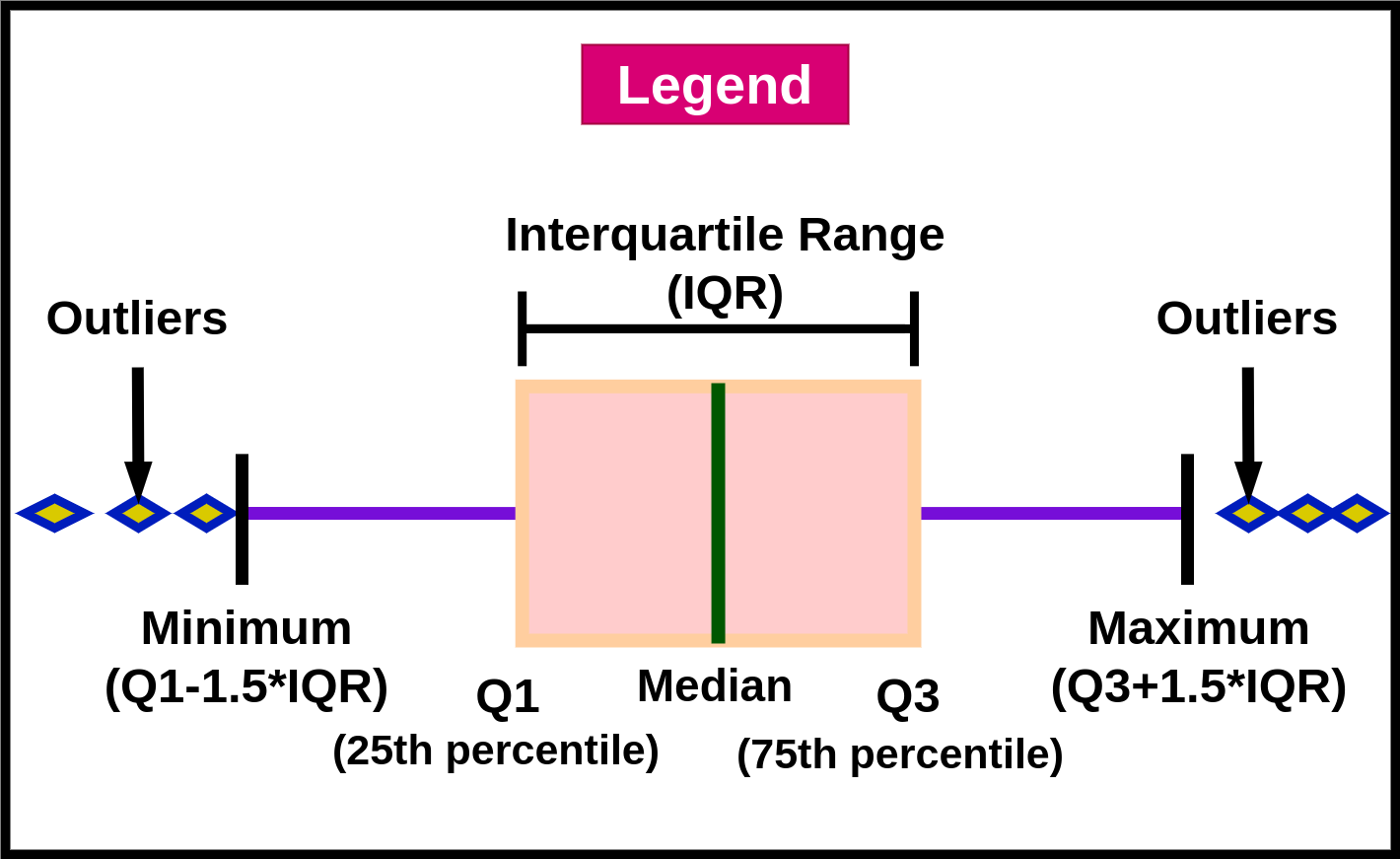}
  \end{minipage}%
  \caption{Each plot shows the Average corner error for the algorithms including MHN, DHN, DLKFM, CLKN, SIFT and ours for a different dataset. The legends used in the plots are also given in the lower right corner of the figure.} % 
    \label{fig:averageCornerErrosAllModels}%
\end{figure}
%%%%%%%%%%%%%%%%%%%%% Results Study %%%%%%%%%%%%%%%%%%%%%%%%%
\begin{figure}[t]
\begin{center}
\includegraphics[width=0.84\linewidth]{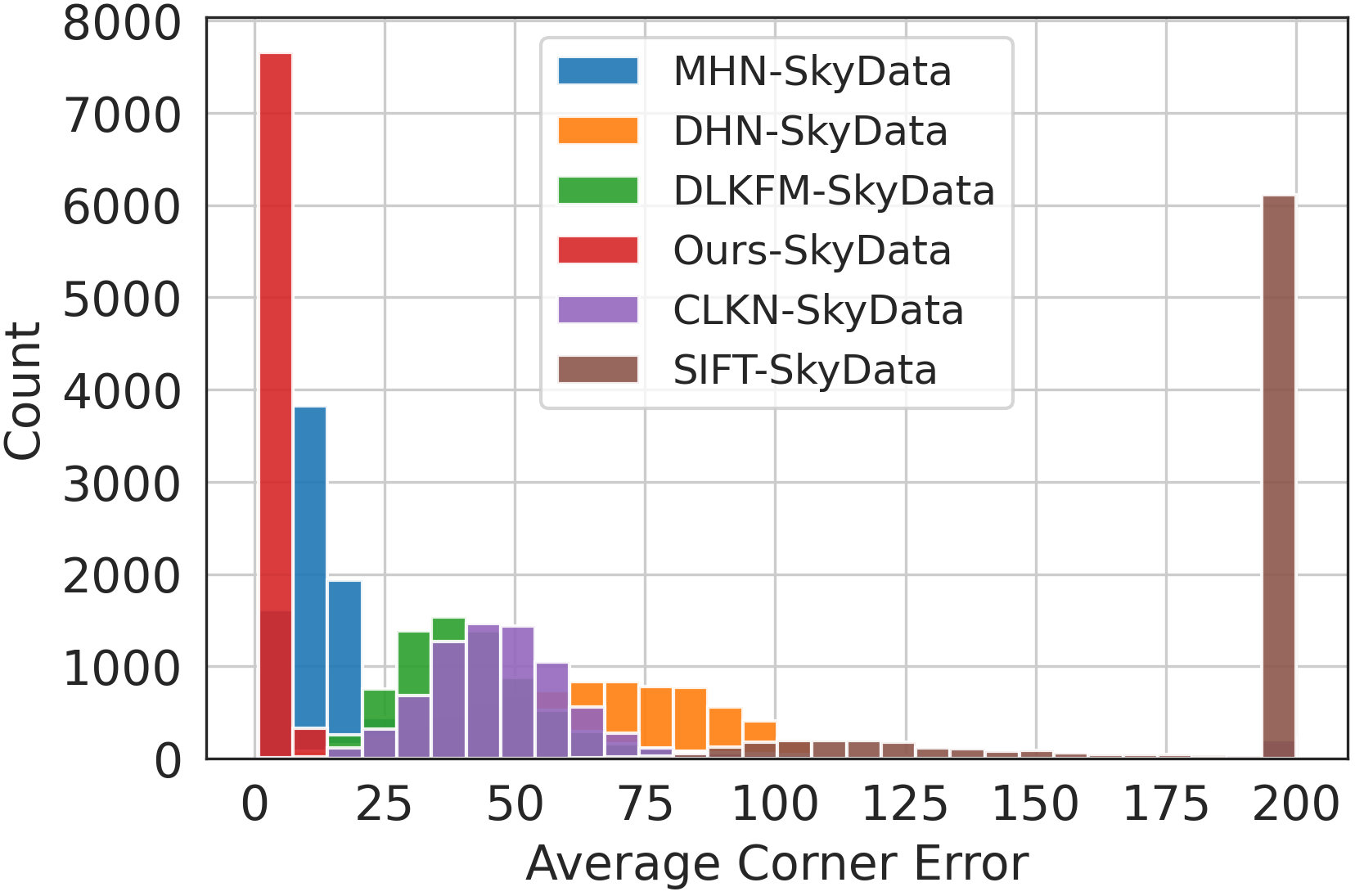}
\end{center}
   \caption{Average corner error distribution vs. count of image pairs for different models is shown on the test set of Skydata.}
\label{fig:AceResults}
\end{figure}

%%%%%% ----------------------- Results tables --------------------------------
\begin{table}[!htb]
    \centering
    \subfloat[t][{\footnotesize  SkyData Results}]{%
        \begin{minipage}[t]{0.5\textwidth}
              \scriptsize      
              \setlength\tabcolsep{.8pt}
              \begin{tabular} {l *{1} {p{1cm} }p{1.2cm} *{4}{p{.9cm}}p{1.2cm}}
                \hline
                    ~ & mean & std & min & 25\% & 50\% & 75\% & max \\ \hline
                    MHN~\cite{Le_2020_CVPR} & 14.5 & 67.29 & 1.19 & 8.11 & 11.36 & 15.4 & 4296.21 \\ \hline
                    DHN~\cite{DBLP:journals/corr/DeToneMR16} & 77.96 & 854.59 & 4.1 & 48.34 & 65.95 & 82.19 & 76119.41 \\ \hline
                    DLKFM~\cite{DBLP:journals/corr/abs-2104-11693} & 93.4 & 2894.7 & 7.32 & 32.08 & 40.73 & 51.95 & 258091.03 \\ \hline
                    Ours & \textbf{3.83} & \textbf{1.77} & \textbf{0.78} & \textbf{2.64} & \textbf{3.46} & \textbf{4.61} & \textbf{19.19} \\ \hline
                    CLKN~\cite{Chang_2017_CVPR} & 77.31 & 862.85 & 5.47 & 38.66 & 47.72 & 57.74 & 73661.8 \\ \hline
                    SIFT~\cite{lowe2004distinctive} & 43477.63 & 201670.37 & 2.13 & 232.88 & 1275.96 & 1285.6 & 100000.0 \\ \hline
                \end{tabular}
        \end{minipage}
    }
    \hfill      
    \subfloat[t][{\footnotesize  VEDAI~\cite{VEDAI} Results}]{%
        \begin{minipage}[t]{0.5\textwidth}
            \scriptsize        
            \setlength\tabcolsep{.8pt}
            \begin{tabular} {l *{1} {p{1cm} }p{1.2cm} *{4}{p{.9cm}}p{1.2cm}}
            \hline
                ~ & mean & std & min & 25\% & 50\% & 75\% & max \\ \hline
                DLKFM~\cite{DBLP:journals/corr/abs-2104-11693} & 382.4 & 3363.72 & 10.93 & 101.43 & 120.09 & 187.6 & 189375.0 \\ \hline
                Ours & \textbf{24.76} & \textbf{4.77} & 4.09 & 21.59 & \textbf{24.88} & \textbf{28.03} & \textbf{42.77} \\ \hline
                MHN~\cite{Le_2020_CVPR} & 374.53 & 5519.56 & 8.74 & 56.4 & 103.68 & 141.99 & 319559.19 \\ \hline
                SIFT~\cite{lowe2004distinctive} & 40221.01 & 195588.26 & \textbf{0.11} & \textbf{2.07} & 88.96 & 1264.86 & 100000.0 \\ \hline
                DHN~\cite{DBLP:journals/corr/DeToneMR16} & 163.87 & 427.16 & 41.59 & 128.8 & 137.43 & 146.43 & 19138.92 \\ \hline
                CLKN~\cite{Chang_2017_CVPR} & 99.49 & 709.1 & 2.45 & 37.97 & 52.29 & 75.13 & 30289.16 \\ \hline
            \end{tabular}
        \end{minipage}
    }
    \hfill
    \subfloat[t][{\footnotesize GoogleEarth Results}]{% 
       \begin{minipage}[t]{0.5\textwidth}
            \scriptsize
            \setlength\tabcolsep{.8pt}
            \begin{tabular} {l *{1} {p{1cm} }p{1.2cm} *{4}{p{.9cm}}p{1.2cm}}
            \hline
                    ~ & mean & std & min & 25\% & 50\% & 75\% & max \\ \hline
                    DHN~\cite{DBLP:journals/corr/DeToneMR16} & 1073.9 & 25214.41 & 8.85 & 52.1 & 67.17 & 83.78 & 733889.5 \\ \hline
                    Ours & \textbf{10.91}& \textbf{3.61}&        4.19&        8.25&         10.5&      \textbf{12.9}&      \textbf{31.13} \\ \hline
                    SIFT~\cite{lowe2004distinctive} & 1334.41 & 34297.53 & 0.51 & \textbf{2.63} & 8.45 & 108.25 & 100000.0 \\ \hline
                    DLKFM~\cite{DBLP:journals/corr/abs-2104-11693} & 27.65 & 121.65 & \textbf{0.36} & 7.41 & 18.5 & 28.96 & 2733.34 \\ \hline
                    MHN~\cite{Le_2020_CVPR} & 118.86 & 257.47 & 12.62 & 45.06 & 60.91 & 106.44 & 4552.5 \\ \hline
                    CLKN~\cite{Chang_2017_CVPR} & 14.58 & 36.57 & 0.38 & 3.15 & \textbf{6.51} & 14.51 & 730.46 \\ \hline
            \end{tabular}
        \end{minipage}
     }  
    \hfill
    \subfloat[h][{\footnotesize GoogleMaps Results}]{%
        \begin{minipage}[t]{0.5\textwidth}
           \scriptsize
            \setlength\tabcolsep{.8pt}
            \begin{tabular} {l *{1} {p{1cm} }p{1.2cm} *{4}{p{.9cm}}p{1.2cm}}
            \hline
                ~ & mean & std & min & 25\% & 50\% & 75\% & max \\ \hline
                MHN~\cite{Le_2020_CVPR} & 319.68 & 1003.02 & 9.48 & 40.68 & 73.54 & 208.8 & 12910.34 \\ \hline
                SIFT~\cite{lowe2004distinctive} & 178912.9 & 382210.28 & 48.79 & 1273.04 & 1281.86 & 1292.6 & 100000.0 \\ \hline
                CLKN~\cite{Chang_2017_CVPR} & 123.96 & 453.65 & 13.22 & 56.43 & 66.9 & 80.77 & 8496.15 \\ \hline
                DHN~\cite{DBLP:journals/corr/DeToneMR16} & 131.27 & 410.38 & 12.28 & 30.27 & 43.08 & 115.07 & 9866.02 \\ \hline
                Ours & \textbf{9.57} & \textbf{4.15} & 2.6 & \textbf{6.82} & \textbf{8.68} & \textbf{11.28} & \textbf{33.67} \\ \hline
                DLKFM~\cite{DBLP:journals/corr/abs-2104-11693} & 77.78 & 183.6 & \textbf{0.47} & 20.27 & 38.76 & 66.28 & 3251.74 \\ \hline
            \end{tabular}
       \end{minipage}
    }
    \hfill
      \subfloat[!h][{\footnotesize MSCOCO\cite{DBLP:journals/corr/LinMBHPRDZ14} Results}]{%
        \scriptsize
        \begin{minipage}[t]{0.5\textwidth}
        \setlength\tabcolsep{.8pt}
        \begin{tabular} {l *{1} {p{1cm} }p{1.2cm} *{4}{p{.9cm}}p{1.2cm}}
        \hline
            ~ & mean & std & min & 25\% & 50\% & 75\% & max \\ \hline
            DLKFM~\cite{DBLP:journals/corr/abs-2104-11693} & 67.16 & 2515.37 & \textbf{0.06} & 0.44 & 8.31 & 31.12 & 200374.44 \\ \hline
            Ours & \textbf{3.67} &\textbf{ 2.45} & 0.64 & 2.28 & 2.99 & 4.13 & \textbf{27.14} \\ \hline
            CLKN~\cite{Chang_2017_CVPR}  & 6.45 & 8.96 & 0.1 & 1.68 & 3.86 & 8.01 & 280.96 \\ \hline
            DHN~\cite{DBLP:journals/corr/DeToneMR16}  & 622.38 & 7493.6 & 3.0 & 141.93 & 194.88 & 383.56 & 580642.19 \\ \hline
            SIFT~\cite{lowe2004distinctive} & 3308.58 & 57236.2 & 0.07 & \textbf{0.37} & \textbf{0.65} & \textbf{1.38} & 100000.0 \\ \hline
            MHN~\cite{Le_2020_CVPR}  & 15.5 & 7.31 & 1.84 & 10.17 & 14.39 & 19.32 & 90.29 \\ \hline
        \end{tabular}
        \end{minipage}
    }
    \caption{\small Comparative results of algorithms on each dataset in terms of Ace. Best results are shown in bold. The results illustrate that on average traditional SIFT performed on average in datasets of single or close modalities. In cases where enough pairs were not found SIFT is unable to estimate homography matrix. We assign a constant 10000.0 as the error. Learning based algorithms DHN and MHN directly predict homography matrix without learning common representation also suffer especially on datasets such as Skydata and Google-maps. This is due to the models being unable to create meaningful correspondences for input and target images as a result of modality difference level. Note that our approach has a small standard deviation as opposed to LK based approaches. LK techniques often significantly deviate from the solution depending on number of iterations they are run and initial parameters they receive.}
    \label{table:resultsTables}
\end{table}
%%%%%%%%%%%%%%%%%%%%% Results Study %%%%%%%%%%%%%%%%%%%%%%%%%
\hfill%

\noindent \textbf{Evaluation metrics}: As shown in Table~\ref{table:resultsTables}, we quantitatively evaluate the performance of our models using Ace and homography error. We compute each algorithm's result distribution in terms of quantiles, mean, standard-deviation and min-max values for a given test set. Quartiles are a set of descriptive statistics which summarize central tendency and variability of data ~\cite{rice2007msd}. Quartiles are a specific type of quantiles that divide the data into four equal parts. The three quartiles are denoted as Q1, Q2 (which is also known as the median), and Q3. The 25\% (Q1), 50\% (Q2) and 75\% (Q3) percentiles indicate that  k\% of the data falls below the $k^{th}$ quartile (the bottom right illustration in Fig.\ref{fig:averageCornerErrosAllModels} also illustrates these terms). To find quartiles;  we first sort elements in data being analysed in ascending order. The first quartile is the number of samples that fall below the dataset size*(1/4) element. Likewise the second quartile is the count of elements that fall below dataset size * (2/4) and the third quartile is dataset size * (3/4) th element in sorted dataset. The samples that fall out of (Q1-1.5 * IQR and Q3+1.5IQR) where IQR is inter-quartile range, are considered outliers. The box plot as in Figure~\ref{fig:averageCornerErrosAllModels} illustrates the above mentioned description visually.
 
Table~\ref{tab:ablationstudy} shows an ablation study on using different loss functions in each block in our architecture. The used metric in the table is Ace and the best values are shown in bold. The loss functions in each row are used to train the MMFEB block includes $\mathcal{L}_{sim}$, $\mathcal{L}_{MAE}$ and $\mathcal{L}_{SSIM}$. The loss functions used for the regression block are $\mathcal{L}_{Ace}$ and ${L}^{H}_{2}$. In the table, the last column shows the average error for both $\mathcal{L}_{Ace}$ and ${L}^{H}_{2}$ (over two datasets including SkyData and VEDAI) for each of the used loss functions in the MMFEB block.

 Next, we provide experimental results on the effect of the hyperparameters that we studied for both ModelA and ModelB.
 Table \ref{tab:l12didnotwork} summarizes those results. In particular, we studied the effect of using different loss functions ($L_1$, $L_2$ and $L_{Ace}$) and using different batch sizes for both models. All the experiments were done on the SkyData set. The best results are shown in bold. Overall, ModelB showed promising results  achieving better results when compared to the ModelA. Therefore, for the rest of our experiments, we kept using ModelB only.

 Fig.\ref{fig:averageCornerErrosAllModels} uses a box plot, also known as a box-and-whisker plot, to display the  distribution of average corner error for different different datasets and for different models. It provides a summary of key statistical measures such as the minimum, first quartile (Q1), median (Q2), third quartile (Q3), and maximum. The length of the box indicates the spread of the middle 50\% of the data. The line inside the box represents the median (Q2). The whiskers extend from the box and represent the variability of the data beyond the quartiles, in our case they represent $Q1-1.5*IQR$ and $Q3+1.5*IQR$. Individual data points that lie outside the whiskers are considered outliers and are plotted with diamonds. The figure compares the results for 6 algorithms on 5 different datasets.

 Fig.\ref{fig:AceResults} shows the performance of 6 methods (SIFT, DHN, MHN, CLKN, DLKFM, Ours) on the SkyDataV1 dataset, in terms of average corner error. Skydata has RGB and infrared image pairs. In this figure, we aim to show that feature based registration techniques such as SIFT perform poorly whereas methods that leverage neural networks and learn representations are superior.

 Fig.\ref{fig:QualitativeResults} gives detailed qualitative results of our experiments. Each row represent a sample taken from a different dataset. The columns represent inputs and results for different approaches. Target is (192x192) (first column), and source (second column) (128x128)  are  input image pairs. Warped (third column), is the ground truth projection of source  to the coordinate system of target image and Registered (fourth column) is the warped image overlayed on the target image as shown. Columns from 5 to 10 shows the registered and overlayed results for SIFT, DHN, MHN, CLKN, DLKFM, and Ours (ModelA) for the given input pair. While almost all algorithms relatively done well on Google Earth pair (which provides similar modalities for both target and source images), when the modalities are significantly different, as in the SkyData, Google Maps and VEDAI pairs, the figure shows that SIFT, CLKN, MHN, DHN and DLKFM algorithms can struggle for aligning them and they may not converge to any useful result near the ground truth (see SIFT and CLKN results), while our approach converges to the ground truth by yielding small ACE error for each of those sample pairs.

 Table \ref {table:resultsTables} illustrates the results of using different approaches for each dataset, separately. in Table \ref {table:resultsTables}(e), the MSCOCO results being a single modality dataset, SIFT perform relatively better but there are cases where the algorithm could not find homography due to insufficient pairs. Google earth in (c) also has RGB image pairs but from different seasons. SIFT algorithm is still able to pick enough salient features therefore the performance is still reasonable. (d) Google maps, (a) SkyData and (b) VEDAI have pairs of  significant modality difference. Deep learning based approaches were able to perform registration often with high number of outliers. Our approach was able to perform registration on both single and multi-modal image pairs, specifically we were able to keep the max error minimum as opposed to LK-based approaches.

\section{Conclusion and Discussion}

In this paper, we introduce a novel image alignment algorithm that we call VisIRNet. VisIRNet has two branches and does not have any stage to compute keypoints. Our experimental results show that our proposed algorithm performs state of the art results, when it is compared to the LK based deep approaches.

Our method's main advantages can be listed as follows:
(a) Number of iterations during inference: Above-mentioned Lucas-Kanade based methods (after the training stage), also iterate a number of times during the inference stage and at each iteration, they try minimizing the loss. However, those methods are not guaranteed to converge to the optimal solution and often number of iterations, chosen as a hyperparameter, is an arbitrary number during the inference stage. Such iterative approaches introduce uncertainty for the processing time, as convergence can happen after the first iteration in some situations and after the last iteration in other situations during inference. Such uncertainty also affects the real time processing of images, as they can introduce varying frame per second values. Our Method uses a single pass during inference with make it more applicable to real time applications.

(b) Dependence on the initial \textbf{H} estimate: In addition to the  above-mentioned difference, the LK-based algorithms require an initial estimate of the  homography matrix and the performance (and number of iterations required for convergence) directly depend on the initial estimate of \textbf{H} and therefore it is typically given as input (hyperparameter). While we also have initialization of the weights in our architecture, we do not need an initial estimate of the homography matrix within the architecture as input.  

Image alignment on image pairs taken by different onboard cameras on UAVs is a challenging and important topic for various applications. When the images to be aligned acquired by different modalities, the classic approaches, such as SIFT and RANSAC combination, can yield insufficient results. Deep learning techniques can be more reliable in such situations as our results demonstrate. LK based deep techniques recently shown promise, however, we demonstrate with our approach (VisIRNet) that without designing any LK based block, and by focusing only on the four corner points, we can sufficiently train deep architectures for image alignment.

%----------------------------------------------------------------------------------------------
\section*{Acknowledgments}
 {\small This paper has been produced benefiting from the 2232 International Fellowship for Outstanding Researchers Program of TÜBİTAK (Project No:118C356). However, the entire responsibility of the paper belongs to the owner of the paper. 
 }

{\small
\bibliographystyle{ieee}
\bibliography{egbib}
}

\par
\section*{Authors' Bio:}
 \textbf{Sedat Ozer} received his M.Sc. degree from Univ. of Massachusetts, Dartmouth and his Ph.D. degree from Rutgers University, NJ. He has worked as a research associate in various institutions including Univ. of Virginia and Massachusetts Institute of Technology. His research interests include pattern analysis, remote sensing, object detection \& segmentation, object tracking, visual data analysis, geometric and explainable AI algorithms and explainable fusion algorithms. As a recipient of TUBITAK's international outstanding research fellow and as an Assistant Professor, he is currently at the department of Computer Science at Ozyegin University.

\par
\par
  \textbf{Alain Patrick Ndigande} received his B.Eng. degree from Kocaeli University, Turkey, in 2022. He is currently a M.Sc. student at Ozyegin University. His current research interests are deep learning, image registration and remote sensing.

\end{document}